%% file: main.tex
\documentclass[10pt,twocolumn,letterpaper]{article}

\usepackage{iccv}
\usepackage{times}
\usepackage{epsfig}
\usepackage{graphicx}
\usepackage{amsmath}
\usepackage{amssymb}
\usepackage{multirow}
\usepackage{hhline}
\usepackage{cancel}
\usepackage{commath}
\usepackage{array}
\usepackage[round, sort, numbers]{natbib}

\newcommand{\comment}[1]{}
\newcommand{\smallsim}{\smallsym{\mathrel}{\sim}}

\makeatletter
\newcommand{\smallsym}[2]{#1{\mathpalette\make@small@sym{#2}}}
\newcommand{\make@small@sym}[2]{%
  \vcenter{\hbox{$\m@th\downgrade@style#1#2$}}%
}
\newcommand{\downgrade@style}[1]{%
  \ifx#1\displaystyle\scriptstyle\else
    \ifx#1\textstyle\scriptstyle\else
      \scriptscriptstyle
  \fi\fi
}
\makeatother

\usepackage[pagebackref=true,breaklinks=true,letterpaper=true,colorlinks,bookmarks=false]{hyperref}

 \iccvfinalcopy % *** Uncomment this line for the final submission

\def\iccvPaperID{2289} % *** Enter the ICCV Paper ID here

\ificcvfinal\fi

\begin{document}
\setcitestyle{square}
%%%%%%%%% TITLE
\title{Rank \& Sort Loss for Object Detection and Instance Segmentation}

\author{%
  Kemal Oksuz, Baris Can Cam, Emre Akbas$^*$, Sinan Kalkan\thanks{Equal contribution for senior authorship.}\\
  Dept. of Computer Engineering, Middle East Technical University,
  Ankara, Turkey \\
  \texttt{\{kemal.oksuz, can.cam, eakbas, skalkan\}@metu.edu.tr} 
}
\comment{
\author{First Author\\
Institution1\\
Institution1 address\\
{\tt\small firstauthor@i1.org}
\and
Second Author\\
Institution2\\
First line of institution2 address\\
{\tt\small secondauthor@i2.org}
}
}
\maketitle
\ificcvfinal\thispagestyle{empty}\fi

%%%%%%%%% ABSTRACT
\begin{abstract}
We propose Rank \& Sort (RS) Loss,  a ranking-based loss function to train deep object detection and instance segmentation methods (i.e. visual detectors). RS Loss supervises the classifier, a sub-network of these methods, to rank each positive above all negatives as well as to sort positives among themselves with respect to (wrt.) their localisation qualities (e.g. Intersection-over-Union - IoU). To tackle the non-differentiable nature of ranking and sorting, we reformulate the incorporation of error-driven update with backpropagation as Identity Update, which enables us to model our novel sorting error among positives. With RS Loss, we significantly simplify training: (i) Thanks to our sorting objective, the positives are prioritized by the classifier without an additional auxiliary head (e.g. for centerness, IoU, mask-IoU), (ii) due to its ranking-based nature, RS Loss is robust to class imbalance, and thus, no sampling heuristic is required, and (iii) we address the multi-task nature of visual detectors using tuning-free task-balancing coefficients. Using RS Loss, we train seven diverse visual detectors only by tuning the learning rate, and  show that it consistently outperforms baselines: e.g. our RS Loss improves (i)  Faster R-CNN by $\smallsim3$ box AP and aLRP Loss (ranking-based baseline) by $\smallsim2$ box AP on COCO dataset, (ii) Mask R-CNN with repeat factor sampling (RFS) by $3.5$ mask AP ($\smallsim7$ AP for rare classes) on LVIS dataset; and also outperforms all counterparts. Code is available at: \url{https://github.com/kemaloksuz/RankSortLoss}.

%, our RS Loss on Mask R-CNN achieves  mask AP on COCO test-dev; outperforming all counterparts. 
\end{abstract}

%%%%%%%%% BODY TEXT
\input{sections/1.Introduction}
\input{sections/2.RelatedWork}

\input{sections/3.IdentityUpdate}
\input{sections/4.RankSortLoss}
\input{sections/5.RankSortforVisualDetection}

\input{sections/6.Experiments}
\input{sections/7.Conclusion}

\textbf{Acknowledgments:} This work was supported by the Scientific and Technological Research Council of Turkey (T\"UB\.{I}TAK) 
% through project called ``Object Detection in Videos with Deep Neural Networks’’ 
(under grants 117E054 and 120E494). We also gratefully acknowledge 
%(i) the support of NVIDIA Corporation with the donation of the Tesla K40 GPU and (ii) 
the computational resources kindly provided by T\"UB\.{I}TAK ULAKBIM High Performance and Grid Computing Center (TRUBA) and Roketsan Missiles Inc. used for this research. Dr. Oksuz is supported by the T\"UB\.{I}TAK 2211-A Scholarship. Dr. Kalkan is supported by the BAGEP Award of the Science Academy, Turkey.
{\small
\bibliographystyle{ieee_fullname}
\bibliography{egbib}
}
\input{sections/8.Appendix}

\end{document}

%% file: sections/1.Introduction.tex
\section{Introduction}
\label{sec:Intro}
\begin{figure}
    \centerline{
        \includegraphics[width=0.48\textwidth]{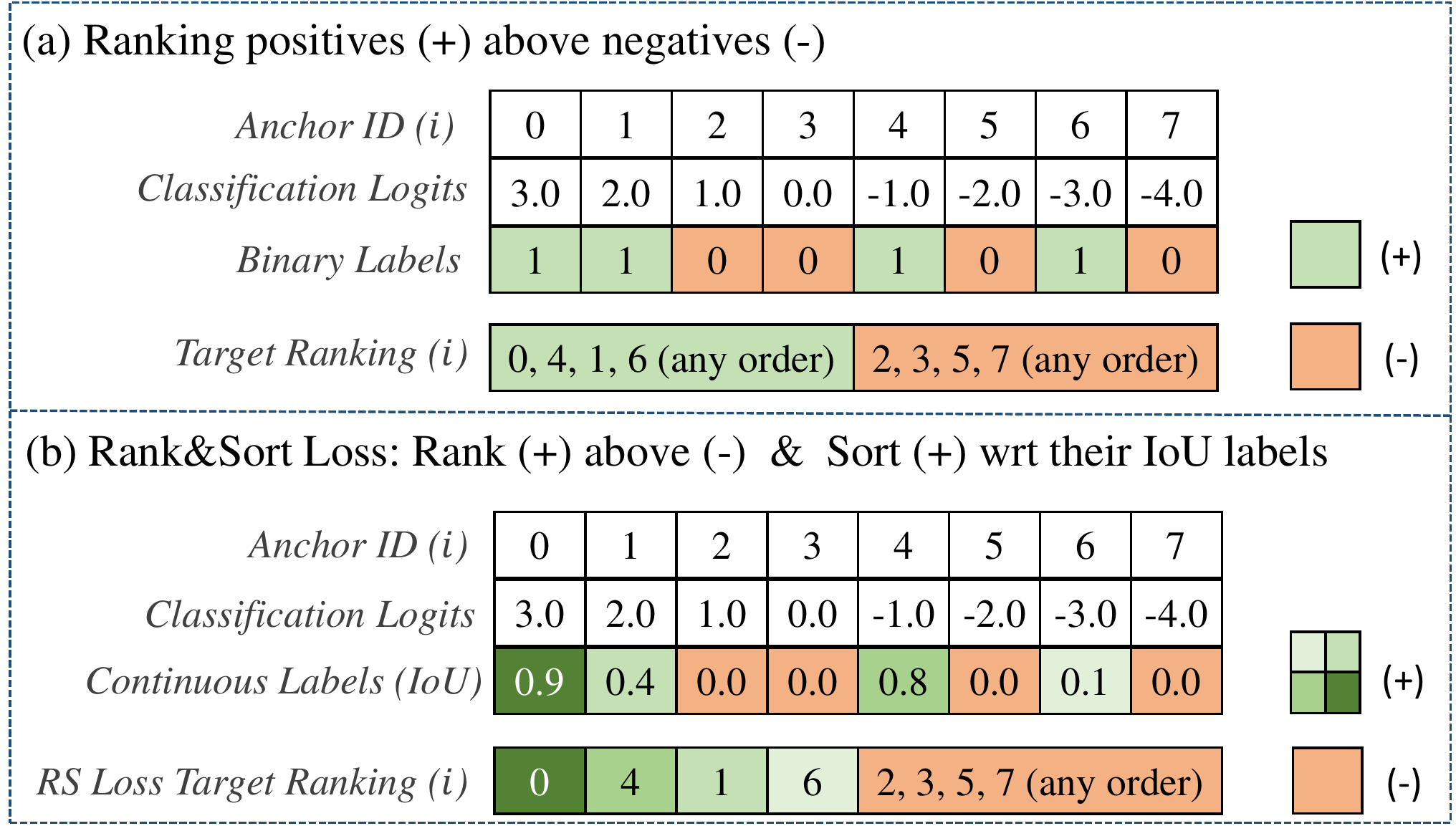}
    }
    \caption{A ranking-based classification loss vs RS Loss. %(a) A generic visual detection pipeline includes many heads from possibly multiple stages. %Mask head is used for instance segmentation and commonly an auxiliary head An auxiliary head, in addition to the standard ones, is common in recent methods (e.g. centerness head for ATSS \cite{ATSS}, IoU head for PAA \cite{paa} and mask IoU head for Mask-scoring R-CNN \cite{maskscoring}). Such heads help learn localization quality and prioritize examples during inference (e.g. by multiplying classification scores by the predicted localisation quality). Sampling heuristics are also common to ensure balanced training. Such architectures require many hyper-parameters and are delicate to tune. 
    (a) Enforcing to rank positives above negatives provides a useful objective for training, however, it ignores ordering among positives. (b) Our RS Loss, in addition to raking positives above negatives, aims to sort positives wrt. their continuous IoUs (positives: a green tone based on its label, negatives: orange).  We propose  Identity Update (Section \ref{sec:IdentityUpdate}), a reformulation of  error-driven update with backpropagation, to tackle these ranking and sorting operations which are difficult to optimize due to their non-differentiable nature. %(both in dashed boxes in (a)) and still outperform baseline networks. 
    %(c) We analyze several design choices and show that RS Loss can be combined by localization losses which are simply balanced based on loss values, requiring no tuning different from lambdas, thereby further simplifying training pipelines. Gradients are computed using bold variables, normal font indicates scalers.
    \label{fig:Teaser}
} 
\end{figure}

Owing to their multi-task (e.g. classification, box regression, mask prediction) nature, object detection and instance segmentation methods rely on  loss functions of the form:
\begin{align}
    \label{eq:LossVisualDetection}
    \mathcal{L}_{VD}= \sum \limits_{k \in \mathcal{K}} \sum \limits_{t \in \mathcal{T}} \lambda_{t}^k \mathcal{L}_{t}^k,
\end{align}
which combines $\mathcal{L}_{t}^k$, the loss function for  task $t$ on stage $k$ (e.g. $|\mathcal{K}|=2$ for Faster R-CNN \cite{FasterRCNN} with RPN and R-CNN), weighted by a hyper-parameter $\lambda_{t}^k$. In such formulations, the number of hyper-parameters can easily exceed 10 \cite{aLRPLoss}, with additional hyper-parameters arising from  task-specific imbalance problems \cite{Review}, e.g. the positive-negative imbalance in the classification task, and if a cascaded architecture  is used (e.g. HTC \cite{HTC} employs $3$ R-CNNs with different $\lambda_{t}^k$). Thus, although such loss functions have led to unprecedented successes, they require tuning, which is time consuming, leads to sub-optimal solutions and makes fair comparison of methods challenging. 

%(ii) Such formulations ignore correlation between the tasks \cite{aLRPLoss}, failing to provide e.g. high-quality localisation for high-quality predictions.

%Tuning hyper-parameters in the loss functions of visual detection tasks (e.g. object detection, instance segmentation) is challenging mainly due to (i) their multi-task nature (e.g. classification, box regression and mask learning tasks in instance segmentation \cite{MaskRCNN,yolact}) and (ii) in some cases, multi-stage design (e.g. four detectors, comprising one RPN \cite{FasterRCNN} and three Fast R-CNNs \cite{FastRCNN}, are trained within Cascade R-CNN \cite{CascadeRCNN}). Furthermore, task-specific imbalance problems \cite{Review}, such as the positive-negative imbalance (a.k.a. class imbalance) in the classification task, can require more hyper-parameters, degrading the severity of this challenge and, in some cases, ending up with more than 10 hyperparameters to be tuned together \cite{aLRPLoss}.

Recently proposed \textit{ranking-based} loss functions, namely ``Average Precision (AP) Loss'' \cite{APLoss} and ``average Localisation Recall Precision (aLRP) Loss'' \cite{aLRPLoss}, offer two important advantages over the classical \textit{score-based} functions (e.g. Cross-entropy Loss and Focal Loss \cite{FocalLoss}): (1) They directly optimize the performance measure (e.g. AP), thereby providing consistency between training and evaluation objectives. This also reduces the number of hyper-parameters as the performance measure (e.g. AP) does not typically have any hyper-parameters. (2) They are robust to class-imbalance due to their ranking-based error definition.  Although these losses have yielded state-of-the-art (SOTA)  performances, they need longer training and more augmentation. %, thus, have not been commonly adopted yet.

Broadly speaking, the ranking-based losses (AP Loss and aLRP Loss) focus on ranking positive examples over negatives, but %they ignore any kind of interaction among positives
they do not explicitly model positive-to-positive interactions. However, there is evidence that it is helpful to prioritize predictions wrt. their localisation qualities by using an auxiliary (aux. - e.g. IoU, centerness) head \cite{IoUNet,FCOS,ATSS,paa} or by supervising the classifier to directly regress IoUs of the predictions without an aux. head (as  shown by Li et al. \cite{GFL} in Quality Focal Loss - QFL).
%thanks to discarding their inconsistency as they are trained only with positives but tested on all examples.

%prioritizing  predictions wrt. their IoUs has  a positive effect on the final performance. Istersen burada "prioritizing  predictions wrt. their IoUs" faydali oldugu eskiden beri biliniyordu \cite{aux head ve IoU regress calismalari} diyebiliriz. 

In this paper, we propose Rank \& Sort (RS) Loss as a ranking-based loss function to train visual detection (VD -- i.e. object detection and instance segmentation) methods. RS Loss not only ranks positives above negatives (Fig. \ref{fig:Teaser}(a)) but also sorts positives among themselves with respect to their continuous IoU values (Fig. \ref{fig:Teaser}(b)). This approach brings in several crucial benefits. Due to the prioritization of positives during training, detectors trained with RS Loss do not need an aux. head, and due to its ranking-based nature, RS Loss can handle extremely imbalanced data (e.g. object detection \cite{Review}) without any sampling heuristics. Besides, except for the learning rate, RS Loss does not need any hyper-parameter tuning thanks to our tuning-free task-balancing coefficients. Owing to this significant simplification of training, we can apply RS Loss to different methods (i.e. multi-stage, one-stage, anchor-based, anchor-free) easily (i.e. \textit{only by tuning the learning rate}) and demonstrate that RS Loss consistently outperforms baselines. 

Our contributions can be summarized as follows: 

\noindent \textbf{(1)} We reformulate the incorporation of error-driven optimization into backpropagation to optimize non-differentiable ranking-based losses as \textit{Identity Update}, which uniquely provides interpretable loss values during training and allows definition of intra-class errors (e.g. the sorting error among positives). 

\noindent \textbf{(2)} We propose \textit{Rank \& Sort Loss} that defines a ranking objective between positives and negatives as well as a sorting objective to prioritize positives wrt. their continuous IoUs. Due to this ranking-based nature, RS Loss can  train  models in the presence of highly imbalanced data. 

\noindent \textbf{(3)} We present the effectiveness of RS Loss on a diverse set of four object detectors and three instance segmentation methods only by tuning the learning rate and without any aux. heads or sampling heuristics on the widely-used COCO and long-tailed LVIS benchmarks: E.g. (i) Our RS-R-CNN improves Faster-CNN by $\smallsim3$ box AP on COCO, (ii) our RS-Mask R-CNN improves repeat factor sampling by $\smallsim3.5$ mask AP ($\smallsim7$ AP for rare classes) on LVIS.

%% file: sections/2.RelatedWork.tex
\section{Related Work}
\textbf{Auxiliary heads and continuous labels.} Predicting the localisation quality of a detection with an aux. centerness \cite{FCOS,ATSS}, IoU \cite{IoUNet,paa}, mask IoU \cite{maskscoring} or uncertainty head \cite{KLLoss} and combining these predictions with the classification scores for NMS are shown to improve detection performance.  Lin et al. \cite{GFL} discovered  that using continuous IoUs of predictions to supervise the classifier outperforms using an aux. head. Currently, Lin et al.'s ``Quality Focal Loss'' \cite{GFL} is the only method that is robust to class imbalance \cite{Review} and uses continuous labels to train the classifier. With RS Loss, we investigate the generalizability of this idea on different networks (e.g. multi-stage networks \cite{FasterRCNN,CascadeRCNN}) and on a different task (i.e. instance segmentation). % by using our ranking-based RS Loss.

\textbf{Ranking-based losses in VD}. Despite their advantages,  ranking-based losses are  non-differentiable and difficult to optimize. To address this challenge, black-box solvers \cite{RankBasedBlackboxDifferentiation} use an interpolated AP surface, though yielding little gain in object detection. DR Loss \cite{DRLoss} achieves ranking between positives and negatives by enforcing a margin with Hinge Loss. Finally, AP Loss \cite{APLoss} and aLRP Loss \cite{aLRPLoss} optimize the performance metrics, AP and LRP \cite{LRP} respectively, by using the error-driven update of perceptron learning \cite{Rosenblatt} for the non-differentiable parts. However, they need longer training and heavy augmentation. The main difference of RS Loss is that it also considers continuous localisation qualities as labels.

\textbf{Objective imbalance in VD}. The common strategy in VD is to use $\lambda_t^k$ (Eq. \ref{eq:LossVisualDetection}), a scalar multiplier, on each task and tune them by grid search \cite{paa,yolact}. Recently, Oksuz et al. \cite{aLRPLoss} employed a self-balancing strategy to balance classification and box regression heads, both of which compete for the bounded range of aLRP Loss. Similarly, Chen et al. \cite{SamplingHeuristics} use the ratio of classification and regression losses to balance these tasks. In our design, each loss $\mathcal{L}_t^k$ for a specific head has its own bounded range and thus, no competition ensues among heads. Besides, we use $\mathcal{L}_t^k$s with similar ranges, and  show that our RS Loss can simply be combined with a simple task balancing strategy based on loss values, and hence does not require any tuning except the learning rate.

%% file: sections/3.IdentityUpdate.tex
\section{Identity Update for Ranking-based Losses}
\label{sec:IdentityUpdate}
Using a ranking-based loss function is attractive thanks to its compatibility with common performance measures (e.g. AP). It is challenging, however, due to the non-differentiable nature of ranking. Here, we first revisit an existing solution \cite{APLoss,aLRPLoss} that overcomes this non-differentiability by incorporating error-driven update \cite{Rosenblatt} into backpropagation (Section \ref{subsec:Errordriven}), and then present our reformulation (Section \ref{subsec:IdentityUpdate}), which uniquely (i) provides interpretable loss values and (ii) takes into account intra-class errors, which is crucial for using continuous labels.

\begin{figure}
    \centerline{
        \includegraphics[width=0.48\textwidth]{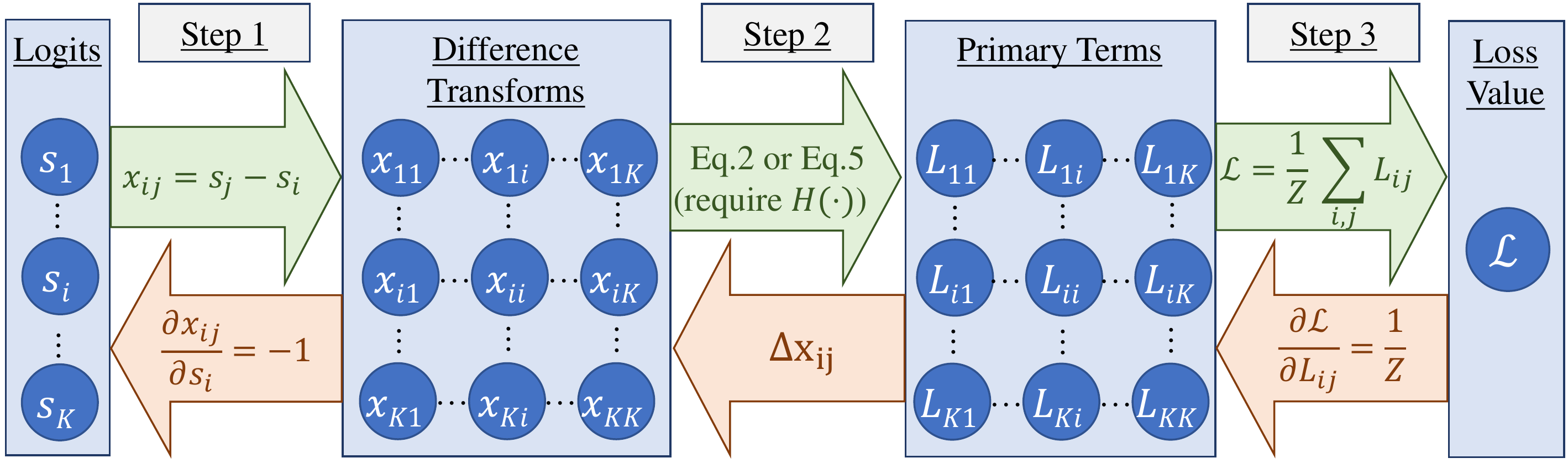}
    }
    \caption{Three-step computation (green arrows) and optimization (orange arrows) algorithms of ranking-based loss functions. Our identity update (i) yields interpretable loss values (see Appendix \ref{sec:RSLossDetails} for an example on our RS Loss), (ii) replaces Eq. \ref{eq:GeneralPrimaryTermDefinition} of previous work \cite{aLRPLoss} by Eq. \ref{eq:PrimaryTermIdentityUpdate} (green arrow in Step 2) to allow intra-class errors, crucial to model our RS Loss, and (iii) results in a simple ``Identity Update'' rule (orange arrow in Step 2): $\Delta x_{ij}=L_{ij}$.
    \label{fig:RankBasedLosses}
} 
\end{figure}

\subsection{Revisiting the Incorporation of Error-Driven Optimization into Backpropagation}
\label{subsec:Errordriven}

\textbf{Definition of the Loss.} Oksuz et al. \cite{aLRPLoss} propose writing a ranking-based loss as $\mathcal{L}= \frac{1}{Z} \sum \limits_{i\in \mathcal{P}} \ell(i)$ where $Z$ is a problem specific normalization constant, $\mathcal{P}$ is the set of positive examples and $\ell(i)$ is the error term computed on $i \in \mathcal{P}$.

\textbf{Computation of the Loss.} Given logits ($s_i$), $\mathcal{L}$ can be computed in three steps \cite{APLoss,aLRPLoss} (Fig. \ref{fig:RankBasedLosses} green arrows):

{\noindent}\textit{Step 1}. The difference transform between logits $s_i$ and $s_j$ is computed by $x_{ij}=s_j-s_i$.

{\noindent}\textit{Step 2}. Using $x_{ij}$, errors originating from each pair of examples are calculated as primary terms ($L_{ij}$):
    \begin{align}
        \label{eq:GeneralPrimaryTermDefinition}
        L_{ij} = \begin{cases} \ell (i) p(j|i), & \mathrm{for}\;i \in \mathcal{P}, j \in \mathcal{N} \\
        0, & \mathrm{otherwise},
        \end{cases}
    \end{align}
where $p(j|i)$ is a probability mass function (pmf) that distributes $\ell (i)$, the error computed on $i \in \mathcal{P}$, over $j \in \mathcal{N}$ where $\mathcal{N}$ is the set of negative examples. By definition, the ranking-based error $\ell (i)$, and thus $L_{ij}$, requires pairwise-binary-ranking relation between outputs $i$ and $j$, which is determined by the non-differentiable unit step function $\mathrm{H}(x)$ (i.e. $\mathrm{H}(x)=1$ if $x \geq 0$ and $\mathrm{H}(x)=0$ otherwise) with input $x_{ij}$. 

Using $\mathrm{H}(x_{ij})$, different ranking-based functions can be introduced to define $\ell (i)$ and $p(j|i)$: e.g. the rank of the $i$th example, $\mathrm{rank}(i)=\sum \limits_{j \in \mathcal{P} \cup \mathcal{N}} \mathrm{H}(x_{ij})$; the rank of the $i$th example among positives, $\mathrm{rank^+}(i)= \sum \limits_{j \in \mathcal{P}} \mathrm{H}(x_{ij})$; and number of false positives with logits larger than $s_i$, $\mathrm{N_{FP}}(i)= \sum \limits_{j \in \mathcal{N}} \mathrm{H}(x_{ij})$. As an example, for AP Loss \cite{APLoss}, using these definitions, $\ell (i)$ and $p(j|i)$ can be simply defined as $\frac{\mathrm{N_{FP}}(i)}{\mathrm{rank}(i)}$ and $\frac{\mathrm{H}(x_{ij})}{\mathrm{N_{FP}}(i)}$ respectively \cite{aLRPLoss}.
    
{\noindent}\textit{Step 3}. Finally, $\mathcal{L}$ is calculated as the normalized sum of the primary terms \cite{aLRPLoss}: $\mathcal{L} =\frac{1}{Z} \sum \limits_{i\in \mathcal{P}} \ell(i)=\frac{1}{Z}\sum \limits_{i \in \mathcal{P} }\sum \limits_{j \in \mathcal{N} }  L_{ij}$.

\textbf{Optimization of the Loss.} Here, the aim is to find updates $\frac{\partial \mathcal{L}}{\partial s_i}$, and then proceed with backpropagation through model parameters. Among the three computation steps (Fig. \ref{fig:RankBasedLosses} orange arrows), Step 1 and Step 3 are differentiable, whereas a primary term $L_{ij}$ is not a differentiable function of difference transforms. Denoting this update in $x_{ij}$ by $\Delta x_{ij}$ and using the chain rule, $\frac{\partial \mathcal{L}}{\partial s_i}$ can be expressed as:
\small
\begin{align}
    \label{eq:Gradients}
    \frac{\partial \mathcal{L}}{\partial s_i} 
    = \sum \limits_{j,k} \frac{\partial \mathcal{L}}{\partial L_{jk}} \Delta x_{jk} \frac{\partial x_{jk}}{\partial s_i} 
    = \frac{1}{Z} \Big( \sum \limits_{j} \Delta x_{ji} - \sum \limits_{j} \Delta x_{ij} \Big). 
\end{align}
\normalsize
Chen et al. \cite{APLoss} incorporate the error-driven update \cite{Rosenblatt} and replace $\Delta x_{ij}$ by $-(L_{ij}^*-L_{ij})$ where $L_{ij}^*$ is the target primary term indicating the desired error for pair $(i, j)$. Both AP Loss \cite{APLoss} and aLRP Loss \cite{aLRPLoss} are optimized this way.%Using this method AP Loss \cite{APLoss} and aLRP Loss \cite{aLRPLoss} are shown to be optimized.

\subsection{Our Reformulation: Identity Update}
\label{subsec:IdentityUpdate}
We first identify two drawbacks of the formulation in Section \ref{subsec:Errordriven}: (D1) Resulting loss value ($\mathcal{L}$) does not consider the target $L_{ij}^*$, and thus, is not easily interpretable when $L_{ij}^* \neq 0$ (cf. aLRP Loss \cite{aLRPLoss} and our RS Loss - Section \ref{sec:RankSortLoss}); (D2) Eq. \ref{eq:GeneralPrimaryTermDefinition} assigns a non-zero primary term only if $i \in \mathcal{P}$ and $j \in \mathcal{N}$, effectively ignoring intra-class errors. These errors become especially important with continuous labels: The larger the label of $i \in \mathcal{P}$, the larger should $s_i$ be. %hence it is limited in addressing intra-class errors, which is crucial, e.g., when the labels are between $0$ and $1$, implying the larger $i \in \mathcal{P}$ has a label, the larger output $s_i$ is to be output by the model; else the ensuing error among positives is to be captured during training, which is not feasible following Eq. \ref{eq:GeneralPrimaryTermDefinition}.

\textbf{Definition of the Loss.} We  redefine the loss function as:
\begin{align}
\label{eq:LossIdentityUpdate}
\mathcal{L}= \frac{1}{Z}\sum \limits_{i \in \mathcal{P} \cup \mathcal{N}} (\ell(i) - \ell^*(i)),    
\end{align}
where $\ell^*(i)$ is the desired error term on $i \in \mathcal{P}$. Our loss definition has two benefits: (i) $\mathcal{L}$ directly measures the difference between the target and the desired errors, yielding an interpretable loss value to address (D1), and %second for the generalization purpose, 
(ii) we do not constrain $\mathcal{L}$ to be defined only on positives and replace ``$i \in \mathcal{P}$" with ``$i \in \mathcal{P} \cup \mathcal{N}$". Although we do not use ``$i \in \mathcal{P} \cup \mathcal{N}$" to model RS Loss, it makes the definition of $\mathcal{L}$ complete in the sense that, if necessary to obtain $\mathcal{L}$, individual errors ($\ell(i)$) can be computed on each output, and hence, $\mathcal{L}$ can be approximated more precisely or a larger set of ranking-based loss functions can be represented.

\textbf{Computation of the Loss.} In order to compute $\mathcal{L}$ (Eq. \ref{eq:LossIdentityUpdate}), we only replace Eq. \ref{eq:GeneralPrimaryTermDefinition} with:
\begin{align}
    \label{eq:PrimaryTermIdentityUpdate}
    L_{ij} = \left(\ell(i) - \ell^*(i) \right) p(j|i),
\end{align}
in three-step algorithm (Section \ref{subsec:Errordriven}, Fig. \ref{fig:RankBasedLosses} green arrows) and allow all pairs to have a non-zero error, addressing (D2).

\textbf{Optimization of the Loss.} Since the error of a pair, $L_{ij}$, is minimized when $\ell(i)=\ell^*(i)$, Eq. \ref{eq:PrimaryTermIdentityUpdate} has a target of $L^*_{ij}=0$ regardless of $\mathcal{L}$. Thus,  $\Delta x_{ij}$ in Eq. \ref{eq:Gradients} is simply the primary term itself: $\Delta x_{ij}=-(L^*_{ij}-L_{ij})=-(0-L_{ij})=L_{ij}$, concluding the derivation of our \textit{Identity Update}.

%% file: sections/4.RankSortLoss.tex
\section{Rank \& Sort Loss}
\label{sec:RankSortLoss}
%Our Rank\&Sort (RS) Loss is different from the existing methods in that not only it is a purely ranking-based loss function, hence robust to class imbalance, but also it supervises object detectors and instance segmentation methods by using continuous localization qualities (e.g. IoU is between $0$ and $1$). Particularly, 
In order to supervise the classifier of visual detectors by considering the localisation qualities of the predictions (e.g. IoU), RS Loss decomposes the problem  into two tasks: (i) \textit{Ranking task}, which aims to rank each positive higher than all negatives, and (ii) \textit{sorting task}, which  aims to sort the logits $s_i$ in descending order wrt. continuous labels $y_i$ (e.g. IoUs). We define RS Loss and compute its gradients using our Identity Update (Section \ref{subsec:IdentityUpdate} -- Fig. \ref{fig:RankBasedLosses}).

\textbf{Definition.} Given logits $s_i$
%\footnote{Similar to previous work \cite{APLoss,aLRPLoss}, our formulation is based on logits instead of probabilities (i.e. scores obtained by applying sigmoid function to these logits). Note that since sigmoid is a strictly increasing function, the order will be preserved in both cases.} 
and their continuous labels $y_i \in [0,1]$ (e.g. IoU), we define RS Loss as the average of the differences between the current ($\ell_{\mathrm{RS}}(i)$) and target ($\ell_{\mathrm{RS}}^*(i)$) RS errors over positives (i.e. $y_i > 0$):
\begin{align}
\label{eq:RSLoss}
    \mathcal{L}_\mathrm{RS}:=\frac{1}{|\mathcal{P}|}\sum \limits_{i \in \mathcal{P}}  \left( \ell_{\mathrm{RS}}(i)-\ell_{\mathrm{RS}}^*(i) \right),
\end{align}
where $\ell_{\mathrm{RS}}(i)$ is a summation of the current ranking error and current sorting error:
\begin{align}
\label{eq:RSCurrent}
    \ell_{\mathrm{RS}}(i):= \underbrace{ \frac{\mathrm{N_{FP}}(i)}{\mathrm{rank}(i)}}_{\ell_{\mathrm{R}}(i):\ \text{Current Ranking Error}}
    +\underbrace{\frac{\sum \limits_{j \in \mathcal{P}} \mathrm{H}(x_{ij})(1-y_j)}{\mathrm{rank^+}(i)}}_{\ell_{\mathrm{S}}(i):\ \text{Current Sorting Error}}. 
\end{align}
For $i \in \mathcal{P}$, while the ``current ranking error'' is simply the precision error, the ``current sorting error''  penalizes the positives with  logits larger than $s_i$ by the average of their inverted labels, $1-y_j$. % of $j \in \mathcal{P}$ with larger logits as a measure to be compared against its target, which is defined based on the target ranking of $i \in \mathcal{P}$ (see target rankings in grey boxes for examples in Fig. \ref{fig:RSLoss}). 
Note that when $i \in \mathcal{P}$ is ranked above all $j \in \mathcal{N}$, $\mathrm{N_{FP}}(i)=0$ and target ranking error, $\ell^*_{\mathrm{R}}(i)$, is $0$. For target sorting error, we average over the inverted labels of $j \in \mathcal{P}$ with larger logits ($\mathrm{H}(x_{ij})$) and labels ($y_j \geq y_i$) than $i \in \mathcal{P}$ corresponding to the desired sorted order,
\begin{align}
\label{eq:RSTarget}
    \ell^*_{\mathrm{RS}}(i)= \cancelto{0}{\ell^*_{\mathrm{R}}(i)}+ \underbrace{\frac{\sum \limits_{j \in \mathcal{P}} \mathrm{H}(x_{ij})[y_j \geq y_i](1-y_j)}{\sum \limits_{j \in \mathcal{P}} \mathrm{H}(x_{ij})[y_j \geq y_i]}}_{\ell^*_{\mathrm{S}}(i):\text{Target Sorting Error}},
\end{align}
where $[\mathrm{P}]$ is the Iverson Bracket (i.e. 1 if predicate $\mathrm{P}$ is True; else 0), and similar to previous work \cite{APLoss}, $\mathrm{H}(x_{ij})$ is smoothed in the interval $[-\delta_{RS},\delta_{RS}]$ as $x_{ij} / 2 \delta_{RS}+0.5$.

\textbf{Computation.} We follow the three-step algorithm (Section \ref{sec:IdentityUpdate}, Fig. \ref{fig:RankBasedLosses}) and define primary terms, $L_{ij}$, using Eq. \ref{eq:PrimaryTermIdentityUpdate}, which allows us to express the errors among positives as:
    \begin{align}\footnotesize
        \label{eq:RSPrimaryTermDefinition}
        L_{ij} = \begin{cases} \left(\ell_{\mathrm{R}}(i) - \ell^*_{\mathrm{R}}(i) \right) p_{R}(j|i), & \mathrm{for}\; i \in \mathcal{P}, j \in \mathcal{N} \\
        \left(\ell_{\mathrm{S}}(i) - \ell^*_{\mathrm{S}}(i) \right) p_{S}(j|i), & \mathrm{for}\;i \in \mathcal{P}, j \in \mathcal{P},\\
        0, & \mathrm{otherwise},
        \end{cases}
    \end{align}
where ranking ($p_{R}(j|i)$) and sorting pmfs ($p_{S}(j|i)$) uniformly distribute ranking and sorting errors on $i$ respectively over examples causing error (i.e. for ranking, $j \in \mathcal{N}$ with $s_j>s_i$; for sorting, $j \in \mathcal{P}$ with $s_j>s_i$ but $y_j < y_i$): 
\begin{align}
    \label{eq:RSpmf}\footnotesize
        p_{R}(j|i) = \frac{\mathrm{H}(x_{ij})}{\sum \limits_{k \in \mathcal{N}}\mathrm{H}(x_{ik})}; p_{S}(j|i) = \frac{\mathrm{H}(x_{ij})[y_j < y_i]}{\sum \limits_{k \in \mathcal{P}} \mathrm{H}(x_{ik})[y_k < y_i]},
\end{align}

\textbf{Optimization.} To obtain $\frac{\partial \mathcal{L}_{\mathrm{RS}}}{\partial s_i}$, we simply replace $\Delta x_{ij}$ (Eq. \ref{eq:Gradients}) by the primary terms of RS Loss, $L_{ij}$ (Eq. \ref{eq:RSPrimaryTermDefinition}), following Identity Update (Section \ref{subsec:IdentityUpdate}). The resulting $\frac{\partial \mathcal{L}_{\mathrm{RS}}}{\partial s_i}$ for $i \in \mathcal{N}$ then becomes  (see Appendix \ref{sec:RSLossDetails} for derivations):
    \begin{align}
    \label{eq:RSNegGradients}
       \frac{\partial \mathcal{L}_{\mathrm{RS}}}{\partial s_i} = \frac{1}{|\mathcal{P}|}\sum \limits_{j \in \mathcal{P}} \ell_{\mathrm{R}}(j)p_{R}(i|j).
    \end{align}
Owing to the additional sorting error (Eq. \ref{eq:RSCurrent}, \ref{eq:RSTarget}), $\frac{\partial \mathcal{L}_{\mathrm{RS}}}{\partial s_i}$ for $i \in \mathcal{P}$ includes update signals for both promotion and demotion to sort the positives accordingly:
\begin{align}
    \label{eq:RSPosGradients}
    \frac{1}{|\mathcal{P}|} \Big( \underbrace{\ell^*_{\mathrm{RS}}(i)-\ell_{\mathrm{RS}}(i)}_{\text{Update signal to promote $i$}} +\underbrace{ \sum \limits_{j \in \mathcal{P} } \left( \ell_{\mathrm{S}}(j)- \ell^*_{\mathrm{S}}(j) \right)  p_{S}(i|j)} _{\text{Update signal to demote $i$}} \Big).
\end{align}
Note that the directions of the first and second part of Eq. \ref{eq:RSPosGradients} are different. To place $i\in \mathcal{P}$ in the desired ranking,  $\ell^*_{\mathrm{RS}}(i)-\ell_{\mathrm{RS}}(i) \leq 0$ promotes $i$ based on the error computed on itself, whereas $  \left( \ell_{\mathrm{S}}(j)- \ell^*_{\mathrm{S}}(j) \right) p_{S}(i|j) \geq 0$ demotes $i$ based on the signal from $j \in \mathcal{P}$. We provide more insight for RS Loss on an example in Appendix \ref{sec:RSLossDetails}.

%% file: sections/5.RankSortforVisualDetection.tex
\section{Using RS Loss to Train Visual Detectors}

\begin{figure}
    \centerline{
        \includegraphics[width=0.45\textwidth]{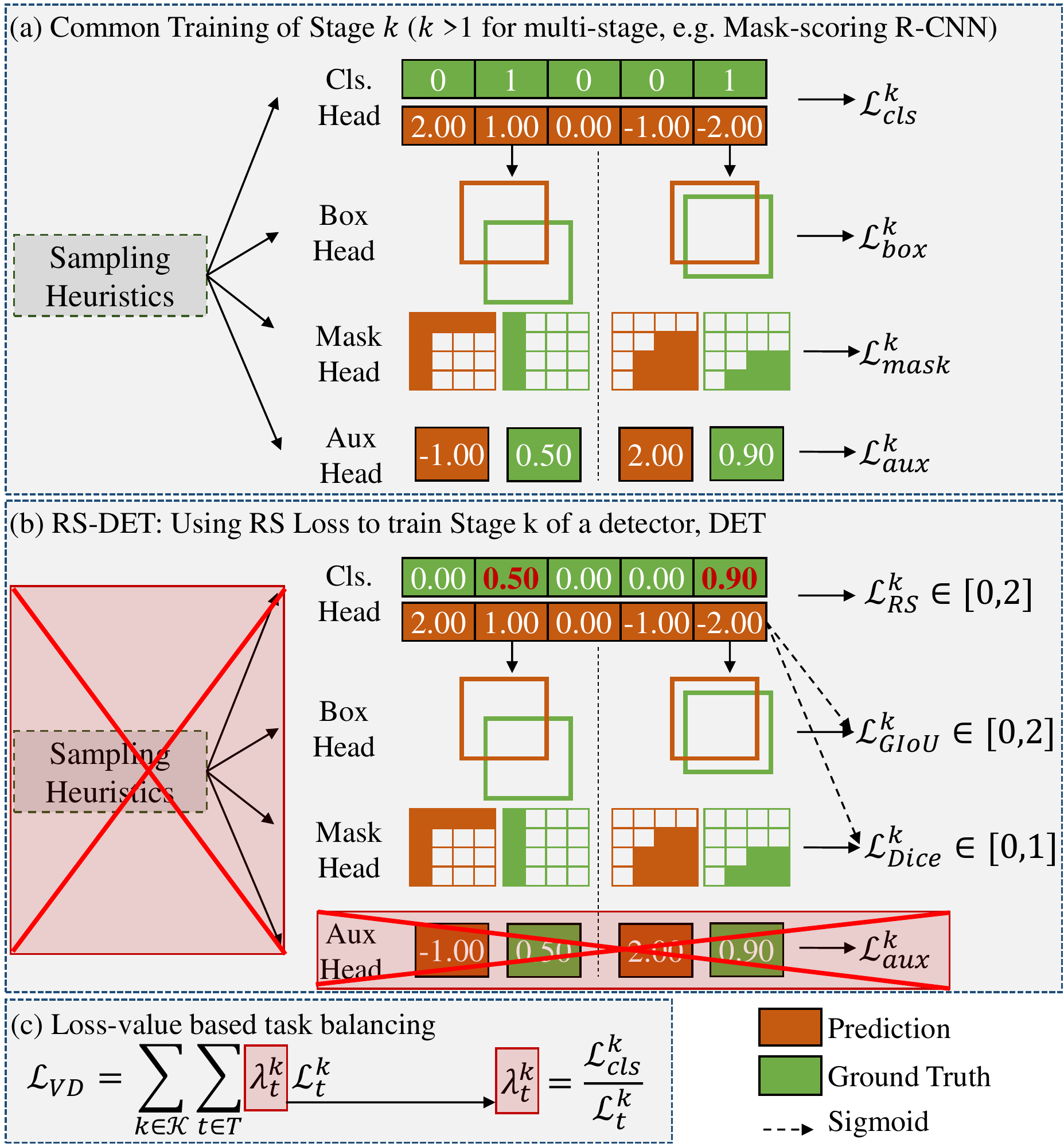}
    }
    \caption{(a) A generic visual detection pipeline includes many heads from possibly multiple stages. An aux. head, in addition to the standard ones, is common in recent methods (e.g. centerness head for ATSS \cite{ATSS}, IoU head for IoU-Net \cite{IoUNet},  and mask IoU head for Mask-scoring R-CNN \cite{maskscoring}) to regress localisation quality and prioritize examples during inference (e.g. by multiplying classification scores by the predicted localisation quality). Sampling heuristics are also common to ensure balanced training. Such architectures use many hyper-parameters and are sensitive for tuning. (b) Training detectors with our RS Loss removes (i) aux. heads by directly supervising the classification (Cls.) head with continuous IoUs (in red \& bold), (ii) sampling heuristics owing to its robustness against class imbalance. We use losses with similar range with our RS Loss in other branches (i.e. GIoU Loss, Dice Loss) also by weighting each by using classification scores, obtained applying sigmoid to logits. (c) Instead of tuning $\lambda_{t}^k$s, we simply balance tasks by considering loss values. With this design, we train several detectors only by tuning the learning rate and improve their performance consistently. 
    \label{fig:Architecture}
} 
\end{figure}
\label{sec:SimpleBaseline}
This section develops an overall loss function to train detectors with RS Loss, in which only the learning rate needs tuning. As commonly performed in the literature \cite{GFL,paa}, Section \ref{subsec:Analysis} analyses different design choices on ATSS \cite{ATSS}, a SOTA one-stage object detector (i.e. $k=1$ in Eq. \ref{eq:LossVisualDetection}); and Section \ref{subsec:Extension} extends our design to other architectures.

\subsection{Dataset and Implementation Details}
Unless explicitly specified, we use (i) standard configuration of each detector and only replace the loss function, (ii) mmdetection framework \cite{mmdetection}, (iii) 16 images with a size of $1333 \times 800$ in a single batch ($4$ images/GPU, Tesla V100) during training, (iv) $1\times$ training schedule (12 epochs),  (v) single-scale test with images with a size of $1333 \times 800$, (vi) ResNet-50 backbone with FPN \cite{FeaturePyramidNetwork}, (vii) COCO \textit{trainval35K} (115K images) and \textit{minival} (5k images) sets \cite{COCO} to train and test our models, (iix) report COCO-style AP.

\subsection{Analysis and Tuning-Free Design Choices}
\label{subsec:Analysis}
ATSS \cite{ATSS} with its classification, box regression and centerness heads is originally trained by minimizing:
\begin{align}
    \label{eq:LossATSS}
    \mathcal{L}_{ATSS}=\mathcal{L}_{cls}+ \lambda_{box} \mathcal{L}_{box} + \lambda_{ctr} \mathcal{L}_{ctr},
\end{align}
where $\mathcal{L}_{cls}$ is Focal Loss \cite{FocalLoss}; $\mathcal{L}_{box}$ is GIoU Loss \cite{GIoULoss}; $\mathcal{L}_{ctr}$ is Cross-entropy Loss with continuous labels to supervise  centerness prediction; and $\lambda_{box}=2$ and $\lambda_{ctr}=1$. We first remove the centerness head and replace $\mathcal{L}_{cls}$ by our RS Loss (Section \ref{sec:RankSortLoss}), $\mathcal{L}_{RS}$, using $\mathrm{IoU}(\hat{b}_i, b_i)$ between a prediction box ($\hat{b}_i$) and ground truth box ($b_i$) as the continuous labels:
\begin{align}
    \label{eq:LossATSSRS}
    \mathcal{L}_{RS-ATSS}=\mathcal{L}_{RS}+ \lambda_{box} \mathcal{L}_{box},
\end{align}
where $\lambda_{box}$, the \textit{task-level balancing coefficient}, is generally set to a constant scalar by grid search. 

Inspired by recent work \cite{aLRPLoss,SamplingHeuristics}, we investigate two tuning-free heuristics to determine  $\lambda_{box}$ every iteration: (i) value-based: $\lambda_{box} = \mathcal{L}_{RS}/\mathcal{L}_{box}$, and (ii) magnitude-based: $\lambda_{box}= \abs{ \frac{{\partial \mathcal{L}}_{\mathrm{RS}}}{{\partial \mathbf{\hat{s}}}}} / \abs{ \frac{{\partial \mathcal{L}}_{\mathrm{box}}}{{\partial \mathbf{b}}}}$ where $\abs{\cdot}$ is L1 norm, $\mathbf{\hat{b}}$ and $\mathbf{s}$ are box regression and classification head outputs respectively. In our analysis on ATSS trained with RS Loss, we observed that value-based task balancing performs similar to tuning $\lambda_{box}$ ($\smallsim 0$ AP difference on average). Also, we use score-based weighting \cite{GFL} by multiplying the GIoU Loss of each prediction using its classification score (details are in Appendix \ref{sec:Analysis}). Note that value-based task balancing and score-based instance weighting are both hyper-parameter-free and easily applicable to all networks. \textit{With these design choices, Eq. \ref{eq:LossATSSRS} has only $1$ hyper-parameter} (i.e. $\delta_{RS}$ in $\mathrm{H}(\cdot)$, set to $0.50$, to smooth the unit-step function)

\subsection{Training Different Architectures}
\label{subsec:Extension}
Fig. \ref{fig:Architecture} presents a comparative overview on how we adopt RS Loss to train different architectures: When we use RS Loss to train the classifier (Fig. \ref{fig:Architecture}(b)), we remove aux. heads (e.g. IoU head in IoU-Net \cite{IoUNet}) and sampling heuristics (e.g. OHEM in YOLACT \cite{yolact}, random sampling in Faster R-CNN \cite{FasterRCNN}). We adopt score-based weighting in box regression and mask prediction heads, and prefer Dice Loss, instead of the common Cross-entropy Loss, to train mask prediction head for instance segmentation due to (i) its bounded range (between $0$ and $1$), and (ii) holistic evaluation of the predictions, both similar to GIoU Loss. Finally, we set $\lambda_{t}^k$ (Eq. \ref{eq:LossVisualDetection}) to scalar $\mathcal{L}_{cls}^k/\mathcal{L}_{t}^k$ (i.e. $\mathcal{L}_{cls} = \mathcal{L}_{RS}$) every iteration (Fig. \ref{fig:Architecture}(c)) with the single exception of RPN where we multiply the losses of RPN by $0.20$ following aLRP Loss.

%% file: sections/6.Experiments.tex
\section{Experiments}
\label{sec:Experiments}
\begin{table*}
    \centering
    \setlength{\tabcolsep}{0.16em}
    \footnotesize
    \caption{RS-R-CNN uses the standard IoU-based assigner, is sampling-free, employs no aux. head, is almost tuning-free wrt. task-balancing weights ($\lambda_{t}^k$s -- Eq. \ref{eq:LossVisualDetection}), and thus, has the least number of hyper-parameters (H\# = $3$ -- two $\delta_{RS}$, one for each RS Loss to train RPN \& R-CNN, and one RPN weight). Still, RS-R-CNN improves standard Faster R-CNN with FPN by $\smallsim3$ AP; aLRP Loss (ranking-based loss baseline) by $\smallsim2$ AP; IoU-Net (a method with IoU head) by $1.5$ AP. RS-R-CNN+ replaces upsampling of FPN by lightweight Carafe operation \cite{carafe} and maintains $\smallsim2$ AP gap from Carafe FPN ($38.6$ to $40.8$ AP). All models use ResNet-50, are evaluated in COCO \textit{minival} and trained for 12 epochs on mmdetection except for IoU-Net. 
    %and KL Loss. Var.: Variance. 
    H\#: Number of hyper-parameters (Appendix \ref{sec:Experiments_supp} presents details on H\#.)}
    \label{tab:TwoStage}
    \begin{tabular}{|c|c|c|c||c|c|c||c|c|c|c||c||c|} \hline
    Method&Assigner&Sampler&Aux. Head&AP $\uparrow$&$\mathrm{AP_{50}} \uparrow$&$\mathrm{AP_{75}} \uparrow$&oLRP $\downarrow$&$ \mathrm{oLRP_{Loc}} \downarrow$&$ \mathrm{oLRP_{FP}} \downarrow$&$ \mathrm{oLRP_{FN}} \downarrow$&H\# $\downarrow$&Venue\\ \hline
    FPN \cite{FeaturePyramidNetwork}& IoU-based&Random&None&$36.5$&$58.5$&$39.4$&$70.1$&$18.3$&$27.8$&$45.8$&$9$&CVPR 17\\
    %OHEM \cite{OHEM}& IoU-based &OHEM& & & & & & & & & CVPR 16\\
    %L1 Loss \cite{mmdetection}& IoU-based&Random&$37.4$&$58.1$&$40.4$&$69.6$&$17.5$&$27.9$&$46.4$& -- (?)\\
    aLRP Loss \cite{aLRPLoss}& IoU-based&None&None&$37.4$&$57.9$&$39.2$&$69.2$&$17.6$&$28.5$&$46.1$ &$\mathbf{3}$&NeurIPS 20\\
    GIoU Loss \cite{GIoULoss}& IoU-based&Random&None&$37.6$&$58.2$&$41.0$&$69.2$&$17.0$&$28.5$&$46.3$&$7$&CVPR 19\\
    %Sparse R-CNN \cite{sparsercnn}&Hungarian&None&$37.9$&$56.0$&$40.5$&$69.2$&$15.8$&$28.6$&$48.5$&CVPR 21\\
    IoU-Net \cite{IoUNet}& IoU-based&Random&IoU Head&$38.1$&$56.3$&--&--&--&--&--&$11$&ECCV 18\\
    Libra R-CNN \cite{LibraRCNN}& IoU-based&IoU-based&None&$38.3$&$59.5$&$41.9$&$68.8$&$17.2$&$27.5$&$45.4$&$11$&CVPR 19\\
    AutoLoss-A \cite{autoloss}& IoU-based&Random&None&$38.5$&$58.6$&$41.8$&$68.4$&$16.6$&$27.1$&$45.5$&$7$&ICLR 21\\
    Carafe FPN \cite{carafe}&IoU-based&Random&None&$38.6$&$59.9$&$42.2$&$68.3$&$17.2$&$27.0$&$44.2$&$7$&ICCV 19\\  
    % KL Loss \cite{KLLoss}& IoU-based&Random&Var. Head&$38.8$&$57.8$&$41.6$&--&--&--&--&$10$&CVPR 19\\
    Dynamic R-CNN \cite{dynamicrcnn}&Dynamic&Random&None&$38.9$&$57.6$&$42.7$&$68.2$&$\mathbf{15.7}$&$27.7$&$46.6$&$10$&ECCV 20\\ \hline 
    RS-R-CNN (Ours)&IoU-based&None&None&$39.6$&$59.5$&$43.0$&$67.9$&$16.3$&$27.8$&$45.4$&$\mathbf{3}$&\\ 
    RS-R-CNN+ (Ours)&IoU-based&None&None&$\mathbf{40.8}$&$\mathbf{61.4}$&$\mathbf{43.8}$&$\mathbf{66.9}$&$16.3$&$\mathbf{26.4}$&$\mathbf{43.7}$&$\mathbf{3}$&\\ \hline 
    \end{tabular}
\end{table*}

To present the contribution of RS Loss in terms of performance and tuning simplicity, we conduct experiments on seven visual detectors with a diverse set of architectures: four object detectors (i.e. Faster R-CNN \cite{FasterRCNN}, Cascade R-CNN \cite{CascadeRCNN}, ATSS \cite{ATSS} and PAA \cite{paa} -- Section \ref{subsec:ObjectDetection}) and three instance segmentation methods (i.e. Mask R-CNN \cite{MaskRCNN}, YOLACT \cite{yolact} and SOLOv2 \cite{solov2} -- Section \ref{subsec:InstanceSegm}). Finally, Section \ref{subsec:Ablation} presents ablation analysis.
%and one keypoint detector (i.e. Keypoint R-CNN \cite{MaskRCNN} -- Section \ref{subsec:KeypointDet}).  

\subsection{Experiments on Object Detection}
\label{subsec:ObjectDetection}
% We train multi- and one-stage detectors with RS Loss.

\subsubsection{Multi-stage Object Detectors}
\label{subsubsec:multistage}
To train Faster R-CNN \cite{FasterRCNN} and Cascade R-CNN \cite{CascadeRCNN} by our RS Loss (i.e. RS-R-CNN), we remove sampling from all stages (i.e. RPN and R-CNNs), use all anchors to train RPN and $m$ top-scoring proposals/image (by default, $m=1000$ for Faster R-CNN and $m=2000$ Cascade R-CNN in mmdetection \cite{mmdetection}), replace softmax classifiers by binary sigmoid classifiers and set the initial learning rate to $0.012$.

%\textbf{Comparison with different R-CNN variants:}
RS Loss reaches $39.6$ AP on a standard Faster R-CNN and outperforms (Table \ref{tab:TwoStage}): (i) FPN \cite{FeaturePyramidNetwork} (Cross-entropy \& Smooth L1 losses) by $3.4$ AP, (ii) aLRP Loss \cite{aLRPLoss}, a SOTA ranking-based baseline, by $2.2$ AP, (iii) IoU-Net \cite{IoUNet} with aux. head by $1.5$ AP and (iv) Dynamic R-CNN, closest counterpart, by $0.7$ AP. We, then, use the lightweight Carafe \cite{carafe} as the  upsampling operation in FPN and obtain $40.8$ AP (RS-R-CNN+), still maintaining $\smallsim2$ AP gap from Carafe FPN \cite{carafe} ($38.6$ AP) and outperforming all methods in all AP- and oLRP-based \cite{LRP,LRParXiv} performance measures except $\mathrm{oLRP_{Loc}}$, which implies that our main contribution is in classification task trained by our RS Loss and there is still room for improvement in the localisation task. RS Loss also improves the stronger baseline Cascade R-CNN \cite{CascadeRCNN} by $1$ AP from $40.3$ AP to $41.3$ AP (Appendix \ref{sec:Experiments_supp} presents detailed results for Cascade R-CNN). %\textbf{Simplification of training:} 
Finally, RS Loss has the least number of hyper-parameters (H\# = $3$, Table \ref{tab:TwoStage}) and does not need  a sampler, an aux. head  or  tuning of $\lambda_{t}^k$s (Eq. \ref{eq:LossVisualDetection}).

\comment{
\begin{table}
    \centering
    \footnotesize
    \caption{\textbf{Improvements to obtain RSVD R-CNN+}}
    \label{tab:Improvements}
    \begin{tabular}{|c|c|c|c||c|c|c|} \hline
    Dyn.Ass.&Carafe&Head+&Cos.Ann.&AP&$\mathrm{AP_{50}}$&$\mathrm{AP_{75}}$\\ \hline 
     & & & &$39.6$&$59.5$&$43.0$ \\ \hline
    \checkmark& & & &$39.4$&$59.5$&$42.6$\\
    &\checkmark&& &$40.8$&$61.4$&$43.8$\\
    & &\checkmark & &$39.6$&$58.6$&$43.0$\\
    & & &\checkmark &$39.3$ &$59.6$&$42.5$\\  \hline
    \checkmark &\checkmark &\checkmark &\checkmark& & &\\  \hline
    \end{tabular}
\end{table}
}

\subsubsection{One-stage Object Detectors}

\begin{table*}
    \centering
    \setlength{\tabcolsep}{0.5em}
    \footnotesize
    \caption{RS Loss has the least number of hyper-parameters (H\#) and outperforms (i) rank-based alternatives significantly, (ii) the default setting with an aux. head (underlined) by $0.6$ AP, (iii) score-based alternative, QFL, especially on PAA. We test unified losses (i.e. a loss considering localisation quality while training classification head) only without aux. head. All models use ResNet-50.}
    \label{tab:OneStage}
    \begin{tabular}{|c|c|c|c||c|c|c||c||c|c|c|c||c|} \hline
    \multirow{2}{*}{Loss Function}&\multirow{2}{*}{Unified}&\multirow{2}{*}{Rank-based}&\multirow{2}{*}{Aux. Head}& \multicolumn{4}{|c|}{ATSS \cite{ATSS}}&\multicolumn{4}{|c|}{PAA \cite{paa}}&\multirow{2}{*}{H\# $\downarrow$}\\ \cline{5-12}
    & & & &AP $\uparrow$&$\mathrm{AP_{50}} \uparrow$&$\mathrm{AP_{75}} \uparrow$&oLRP $\downarrow$&AP $\uparrow$&$\mathrm{AP_{50}} \uparrow$&$\mathrm{AP_{75}} \uparrow$&oLRP $\downarrow$ &\\ \hline
    \multirow{2}{*}{Focal Loss \cite{FocalLoss}}& & & &$38.7$&$57.6$&$41.5$&$68.9$ &$39.9$&$57.3$&$43.4$&$68.7$&$3$\\
    & & &\checkmark&\underline{$39.3$}&\underline{$57.5$}&\underline{$42.6$}&\underline{$68.6$} &\underline{$40.4$}&\underline{$58.4$}&\underline{$43.9$}&\underline{$67.7$}&\underline{$4$}\\ \hline
    \multirow{2}{*}{AP Loss \cite{APLoss}}& &\checkmark& &$38.1$&$58.2$&$41.0$&$69.2$ &$35.3$&$53.1$&$38.5$&$71.5$&$2$\\
    & &\checkmark&\checkmark&$37.2$&$55.6$&$40.2$&$70.0$&$37.3$&$54.3$&$41.2$&$70.5$&$3$\\ \hline
    QFL \cite{GFL} &\checkmark& & &$39.7$&$58.1$&$\mathbf{42.7}$&$68.0$&$40.2$&$57.4$&$43.8$&$68.3$&$2$\\ \hline
    aLRP Loss \cite{aLRPLoss}&\checkmark&\checkmark& &$37.7$&$57.4$&$39.9$&$69.4$ &$37.7$&$56.1$&$40.1$&$69.9$&$\mathbf{1}$\\ \hline
    RS Loss (Ours) &\checkmark&\checkmark& &$\mathbf{39.9}$&$\mathbf{58.9}$&$42.6$&$\textbf{67.9}$&$\mathbf{41.0}$&$\mathbf{59.1}$&$\mathbf{44.5}$&$\textbf{67.3}$&$\mathbf{1}$\\ \hline
    \end{tabular}
\end{table*}
We train ATSS \cite{ATSS} and PAA \cite{paa} including a centerness head and an IoU head respectively in their architectures. We adopt the anchor configuration of Oksuz et al. \cite{aLRPLoss} for all ranking-based losses (different anchor configurations do not affect performance of standard ATSS \cite{ATSS}) and set learning rate to $0.008$. While training PAA, we keep the scoring function, splitting positives and negatives, for a fair comparison among different loss functions.

\textbf{Comparison with AP and aLRP Losses, ranking-based baselines:} We simply replaced Focal Loss by AP Loss to train networks, and as for aLRP Loss, similar to our RS Loss, we tuned its learning rate as $0.005$ due to its tuning simplicity. Both for ATSS and PAA, RS Loss provides significant gains over ranking-based alternatives, which were trained for 100 epochs using SSD-like augmentation \cite{SSD} in previous work \cite{APLoss,aLRPLoss}: $1.8$/$2.2$ AP gain for ATSS and $3.7$/$3.3$ gain for PAA for AP/aLRP Loss (Table \ref{tab:OneStage}).
 
\textbf{Comparison with Focal Loss, default loss function:} RS Loss provides around $\smallsim 1$ AP gain when both networks are equally trained without an aux. head (Table \ref{tab:OneStage}) and $0.6$ AP gain compared to the default networks with aux. heads.

\textbf{Comparison with QFL, score-based loss function using continuous IoUs as labels:} To apply QFL \cite{GFL} to PAA,  we remove the aux. IoU head (as we did with ATSS), test two possible options ((i) default PAA setting with $\lambda_{box}=1.3$ and IoU-based weighting, (ii) default QFL setting: $\lambda_{box}=2.0$ and score-based weighting) and report the best result for QFL. While the results of QFL and RS Loss are similar for ATSS, there is $0.8$ AP gap in favor of our RS Loss, which can be due to the different positive-negative assignment method of PAA (Table \ref{tab:OneStage}). 

\subsubsection{Comparison with SOTA}
Here, we use our RS-R-CNN since it yields the largest improvement over its baseline. %for which our performance improvement is more significant against baseline.
We train RS-R-CNN for 36 epochs using multiscale training by randomly resizing the shorter size within $[480, 960]$ on ResNet-101 with DCNv2 \cite{DCNv2}. Table \ref{tab:Detectiontestdev} reports the results on COCO \textit{test-dev}: Our RS-R-CNN reaches $47.8$ AP and outperforms similarly trained Faster R-CNN and Dynamic R-CNN by $\smallsim 3$ and $\smallsim 1$ AP respectively. Although we do not increase the number of parameters for Faster R-CNN, RS R-CNN outperforms all multi-stage detectors including TridentNet \cite{TridentNet}, which has more parameters. Our RS-R-CNN+ (Section \ref{subsubsec:multistage}) reaches $48.2$ AP, and RS-Mask R-CNN+ (Section \ref{subsec:InstanceSegm}) reaches $49.0$ AP, outperforming all one- and multi-stage counterparts.

\begin{table}
    \centering
    \setlength{\tabcolsep}{0.25em}
    \footnotesize
    \caption{Comparison with SOTA for object detection on COCO \textit{test-dev} using ResNet-101 (except *) with DCN. The result of the similarly trained Faster R-CNN is acquired from Zhang et al. \cite{dynamicrcnn}.  +: upsampling of FPN is Carafe \cite{carafe}, *:ResNeXt-64x4d-101}
    \label{tab:Detectiontestdev}
    \begin{tabular}{|m{0.8em}|c||c|c|c|c|c|c|} \hline
    &Method&AP&$\mathrm{AP_{50}}$&$\mathrm{AP_{75}}$&$\mathrm{AP_{S}}$&$\mathrm{AP_{M}}$&$\mathrm{AP_{L}}$\\ \hline 
    \multirow{4}{*}{\rotatebox{90}{One-stage}}&ATSS \cite{ATSS}&$46.3$&$64.7$&$50.4$&$27.7$&$49.8$&$58.4$\\
    &GFL \cite{GFL}&$47.3$&$66.3$&$51.4$&$28.0$&$51.1$&$59.2$\\
    &PAA \cite{paa}&$47.4$&$65.7$&$51.6$&$27.9$&$51.3$&$60.6$\\
    &RepPointsv2 \cite{reppointsv2}&$48.1$&$67.5$&$51.8$&$28.7$&$50.9$&$60.8$\\ 
    % &GFLv2 \cite{GFLv2}&$48.3$&$66.5$&$52.8$&$28.8$&$51.9$&$60.7$\\ 
    \hline
    \multirow{4}{*}{\rotatebox{90}{Multi-stage}}&Faster R-CNN \cite{dynamicrcnn}&$44.8$&$65.5$&$48.8$&$26.2$&$47.6$&$58.1$\\
    &Trident Net \cite{TridentNet}&$46.8$&$67.6$&$51.5$&$28.0$&$51.2$&$60.5$ \\
    &Dynamic R-CNN \cite{dynamicrcnn}&$46.9$&$65.9$&$51.3$&$28.1$&$49.6$&$60.0$ \\
    &D2Det \cite{D2Det}&$47.4$&$65.9$&$51.7$&$27.2$&$50.4$&$61.3$\\ \hline
    \multirow{4}{*}{\rotatebox{90}{Ours}}&RS-R-CNN &$47.8$&$68.0$&$51.8$&$28.5$&$51.1$&$61.6$\\ 
    &RS-R-CNN+&$48.2$&$68.6$&$52.4$&$29.0$&$51.3$&$61.7$\\  
    &RS-Mask R-CNN+&$\mathbf{49.0}$&$\mathbf{69.2}$&$\mathbf{53.4}$&$\mathbf{29.9}$&$\mathbf{52.4}$&$\mathbf{62.8}$\\ \cline{2-8}
    &RS-Mask R-CNN+*&$50.2$&$70.3$&$54.8$&$31.5$&$53.5$&$63.9$\\ 
    %\cline{2-8} &RS-Mask R-CNN+$^*$&$50.4$&$70.6$&$55.1$&$32.9$&$53.8$&$62.0$\\  
    \hline
    \end{tabular}
\end{table}

\subsection{Experiments on Instance Segmentation}
\label{subsec:InstanceSegm}
%Similar to Section \ref{subsec:ObjectDetection}, we train multi- and one-stage methods with RS Loss and compare our results with SOTA.
%We train multi- and one-stage methods with RS Loss.

\subsubsection{Multi-stage Instance Segmentation Methods}
We train Mask R-CNN \cite{MaskRCNN} on COCO and LVIS datasets by keeping all design choices of Faster R-CNN the same.

\textbf{COCO:} We observe $\smallsim2$ AP gain for both segmentation and detection performance (Table \ref{tab:MaskRCNN}) over Mask R-CNN. Also, RS-Mask R-CNN outperforms Mask-scoring R-CNN \cite{maskscoring}, with an additional aux. mask IoU head, by $0.4$ mask AP, $1.8$ box AP and $0.9$ mask oLRP (Table \ref{tab:MaskRCNN}). 
\begin{table}
    \centering
    \footnotesize
    \setlength{\tabcolsep}{0.1em}
    \caption{Without an aux. head, RS-Mask R-CNN improves Mask R-CNN  \cite{MaskRCNN} by $\smallsim2$ AP and outperforms Mask-scoring R-CNN \cite{maskscoring} which employs an additional mask IoU head as aux. head.}
    \label{tab:MaskRCNN}
    \begin{tabular}{|c|c||c|c|c|c||c||c|} \hline
    \multirow{2}{*}{Method}&Aux& \multicolumn{4}{|c||}{Segmentation Performance}&\multirow{2}{*}{$\mathrm{AP_{box}}$}&\multirow{2}{*}{H\# $\downarrow$}\\ \cline{3-6}
    &Head&AP $\uparrow$&$\mathrm{AP_{50}} \uparrow$&$\mathrm{AP_{75}} \uparrow$&oLRP $\downarrow$& &\\ \hline
    Mask R-CNN& &$34.7$&$55.7$&$37.2$&$71.2$&$38.2$&$8$\\ 
%    Cascade Mask R-CNN \cite{MaskRCNN}&$35.9$&$56.6$&$38.4$& &$41.2$&$59.4$&$45.0$& \\ 
    Mask-sc. R-CNN &\checkmark&$36.0$&$55.8$&$38.7$& $71.0$&$38.2$&$9$ \\ \hline
    RS-Mask R-CNN& &$\mathbf{36.4}$&$\mathbf{57.3}$&$\mathbf{39.2}$&$\mathbf{70.1}$ &$\mathbf{40.0}$&$\mathbf{3}$ \\  \hline
%    RS Cascade Mask R-CNN &$?$&$?$&$?$& &$?$&$?$&$?$& \\  
%    RS-Mask R-CNN +&$37.3$&$58.6$&$40.2$&$69.4$&$41.1$&$61.4$&$44.9$&$66.6$ \\ \hline
%    RS Cascade Mask R-CNN+&$?$&$?$&$?$& &$?$&$?$&$?$& \\ \hline
    \end{tabular}
\end{table}

\textbf{LVIS:} Replacing the Cross-entropy to train Mask R-CNN with repeat factor sampling (RFS) by our RS Loss improves the performance by $3.5$ mask AP on the long-tailed LVIS dataset ($21.7$ to $25.2$ with $\smallsim 7 \mathrm{AP_{r}}$ improvement on rare classes) and outperforms recent counterparts (Table \ref{tbl:LVIS}). 

\begin{table}
\setlength{\tabcolsep}{0.25em}
\footnotesize
\caption{Comparison on LVIS v1.0 val set. Models are trained with ResNet-50, multiscale images (range: $[640,800]$) for 12 epochs. 
%In addition, BAGS is finetuned for 12 epochs. 
\label{tbl:LVIS}} 
 \centering
\begin{tabular}{|c||c|c|c|c||c||c|}
\hline
Method&$\mathrm{AP_{mask}}$&$\mathrm{AP_{r}}$&$\mathrm{AP_{c}}$&$\mathrm{AP_{f}}$&$\mathrm{AP_{box}}$&Venue\\ \hline
RFS \cite{LVIS} &$21.7$&$9.6$&$21.0$&$27.8$&$22.5$&CVPR 19\\
BAGS \cite{BAGS}&$23.1$&$13.1$&$22.5$&$28.2$&$23.7$&CVPR 20 \\ 
Eq. Lossv2 \cite{Eqv2}&$23.7$&$14.9$&$22.8$&$28.6$&$24.2$&CVPR 21\\ \hline
RFS+RS Loss&$\mathbf{25.2}$&$\mathbf{16.8}$&$\mathbf{24.3}$&$\mathbf{29.9}$&$\mathbf{25.9}$& \\ \hline
\end{tabular}
\end{table}

\subsubsection{One-stage Instance Segmentation Methods}
Here, we train two different approaches with our RS Loss: (i) YOLACT \cite{yolact},  a real-time instance segmentation method, involving sampling heuristics (e.g. OHEM \cite{OHEM}), aux. head and carefully-tuned loss weights, and demonstrate RS Loss can discard all by improving its performance (ii) SOLOv2 \cite{solov2} as an anchor-free SOTA method.

\begin{table*}
    \centering
    \setlength{\tabcolsep}{0.25em}
    \footnotesize
    \caption{RS-YOLACT does not employ any additional training heuristics and outperforms  YOLACT by significant margin.}
    \label{tab:Yolact}
    \begin{tabular}{|c|c|c|c||c|c|c||c||c|c|c|c||c|} \hline
    \multirow{2}{*}{Method}& \multicolumn{3}{|c||}{Additional Training Heuristics}& \multicolumn{4}{|c||}{Segmentation Performance}&\multicolumn{4}{|c|}{Detection Performance}&\multirow{2}{*}{H\# $\downarrow$}\\ \cline{2-12}
    &OHEM \cite{OHEM}&Size-based Norm.&Sem.Segm. Head&AP $\uparrow$&$\mathrm{AP_{50}} \uparrow$&$\mathrm{AP_{75}} \uparrow$&oLRP $\downarrow$&AP $\uparrow$&$\mathrm{AP_{50}} \uparrow$&$\mathrm{AP_{75}} \uparrow$&oLRP $\downarrow$& \\ \hline
    \multirow{4}{*}{YOLACT \cite{yolact}}& \checkmark &\checkmark & \checkmark&$28.4$&$47.7$&$29.1$&$75.4$&$30.5$&$52.3$&$31.8$&$73.9$&$5$ \\
    & &\checkmark&\checkmark&$15.1$&$27.1$&$15.0$&$86.6$&$12.6$&$27.8$&$10.0$&$88.5$&$4$ \\
    &\checkmark& &\checkmark&$21.7$&$40.2$&$20.7$&$80.5$&$30.4$&$52.2$&$31.4$&$74.1$&$5$ \\
    &\checkmark&\checkmark& &$28.1$&$47.5$&$28.7$&$75.6$&$30.5$&$51.9$&$32.0$&$74.0$&$4$ \\
    & & & &$13.6$&$26.4$&$12.4$&$87.5$&$15.1$&$32.9$&$12.1$&$86.3$&$3$ \\ \hline
    %aLRP Loss& & & &$28.6$&$48.6$&$29.0$&$75.2$&$31.9$&$52.8$&$32.4$&$72.9$ \\ \hline
    RS-YOLACT& & & &$\mathbf{29.9}$&$\mathbf{50.5}$&$\mathbf{30.6}$& $\mathbf{74.7}$&$\mathbf{33.8}$&$\mathbf{54.2}$&$\mathbf{35.4}$&$\mathbf{71.8}$&$\mathbf{1}$ \\ \hline
    \end{tabular}
\end{table*}

\textbf{YOLACT:} Following YOLACT \cite{yolact}, we train and test RS-YOLACT by images with size $550 \times 550$ for $55$ epochs. Instead of searching for epochs to decay learning rate, carefully tuned for YOLACT as $20$, $42$, $49$ and $52$, we simply adopt cosine annealing with an initial learning rate of $0.006$. Then, we remove (i) OHEM, (ii) semantic segmentation head, (iii) carefully tuned task weights (i.e. $\lambda_{box}=1.5$, $\lambda_{mask}=6.125$) and (iv) size-based normalization (i.e. normalization of mask head loss of each instance by the ground-truth area). Removing each heuristic ensues a slight to significant performance drop (at least requires retuning of $\lambda_{t}$ -- Table \ref{tab:Yolact}). After these simplifications, our RS-YOLACT improves baseline by $1.5$ mask AP and $3.3$ box AP.
%, and (ii) ranking-based baseline, aLRP Loss, by $1.3$ mask AP and $1.9$ box AP.

\textbf{SOLOv2:} Following Wang et al. \cite{solov2}, we train anchor-free SOLOv2 with RS Loss for 36 epochs using multiscale training on its two different settings: (i) SOLOv2-light is the real-time setting with ResNet-34 and images with size $448 \times 448$ at inference. We use 32 images/batch and learning rate $0.012$ for training. (ii)  SOLOv2 is the SOTA setting with ResNet-101 and images with size $1333 \times 800$ at inference. We use 16 images/batch and learning rate $0.006$ for training. Since SOLOv2 does not have a box regression head, we use Dice coefficient as the continuous labels of RS Loss (see Appendix \ref{sec:Experiments_supp} for an analysis of using different localisation qualities as labels). Again, RS Loss performs better than the baseline (i.e. Focal Loss and Dice Loss) only by tuning the learning rate (Table \ref{tab:SOLOv2}).

\subsubsection{Comparison with SOTA}
We use our RS-Mask R-CNN (i.e. standard Mask R-CNN with RS Loss) to compare with SOTA methods. In order to fit in 16GB memory of our V100 GPUs and keep all settings unchanged, we limit the number of maximum proposals in the mask head by 200, which can simply be omitted for GPUs with larger memory. Following our counterparts \cite{solov2,centermask}, we first train RS-Mask R-CNN for 36 epochs with multiscale training between $[640,800]$ using ResNet-101 and reach $40.6$ mask AP (Table \ref{tab:Segmentationtestdev}), improving Mask R-CNN by $2.3$ mask AP and outperforming all SOTA methods by a notable margin ($\smallsim 1$ AP). Then, we train RS-Mask R-CNN+ (i.e. standard Mask R-CNN except upsampling of FPN is lightweight Carafe \cite{carafe}) also by extending the multiscale range to $[480,960]$ and reach $42.0$ mask AP, which even outperforms all models with DCN. With DCN \cite{DCNv2} on ResNet-101, our RS-Mask R-CNN+ reaches $43.9$ mask AP.

\begin{table}
    \centering
    \footnotesize
    \setlength{\tabcolsep}{0.2em}
    \caption{Comparison on anchor-free SOLOv2.}
    \label{tab:SOLOv2}
    \begin{tabular}{|c|c||c|c|c||c||c|} \hline
    Method&Backbone&AP&$\mathrm{AP_{50}}$&$\mathrm{AP_{75}}$&oLRP $\downarrow$&H\# $\downarrow$ \\ \hline 
    SOLOv2-light&ResNet-34&$32.0$&$50.7$&$33.7$&$73.5$&$3$\\ 
    RS-SOLOv2-light&ResNet-34&$\mathbf{32.6}$&$\mathbf{51.7}$&$\mathbf{34.2}$&$\mathbf{72.7}$&$\mathbf{1}$\\ \hline
    SOLOv2&ResNet-101&$39.1$&$59.8$&$41.9$&$67.3$&$3$\\ 
    RS-SOLOv2&ResNet-101&$\mathbf{39.7}$&$\mathbf{60.6}$&$\mathbf{42.2}$&$\mathbf{66.9}$&$\mathbf{1}$\\ \hline
    \end{tabular}
\end{table}

\begin{table}
    \centering
    \footnotesize
    \setlength{\tabcolsep}{0.15em}
    \caption{Comparison with SOTA for instance segmentation on COCO \textit{test-dev}. All methods (except *) use ResNet-101. The result of the similarly trained Mask R-CNN is acquired from Chen et al. \cite{tensormask}. +: upsampling of FPN is Carafe, *:ResNeXt-64x4d-101}
    \label{tab:Segmentationtestdev}
    \begin{tabular}{|m{0.8em}|c||c|c|c|c|c|c|} \hline
    &Method&AP&$\mathrm{AP_{50}}$&$\mathrm{AP_{75}}$&$\mathrm{AP_{S}}$ &$\mathrm{AP_{M}}$&$\mathrm{AP_{L}}$\\ \hline 
    \multirow{6}{*}{\rotatebox{90}{w/o DCN}}&Polar Mask \cite{polarmask}&$32.1$&$53.7$&$33.1$&$14.7$&$33.8$&$45.3$\\ 
%    D-SOLO \cite{solo}&$38.4$&$59.6$&$41.1$&$16.8$&$41.5$&$54.6$\\ 
%    &Blend Mask \cite{blendmask}&$39.6$&$\mathbf{61.6}$&$42.6$&$\mathbf{22.4}$&$42.2$&$51.4$\\
    &Mask R-CNN \cite{tensormask} &$38.3$&$61.2$&$40.8$&$18.2$&$40.6$&$54.1$ \\
    &SOLOv2 \cite{solov2}&$39.7$&$60.7$&$42.9$&$17.3$&$42.9$&$\mathbf{57.4}$\\
    &Center Mask \cite{centermask} &$39.8$&--&--&$21.7$&$42.5$&$52.0$\\  
    &BCNet \cite{BCNet}&$39.8$&$61.5$&$43.1$&$22.7$&$42.4$&$51.1$\\ \cline{2-8}
%    &RS-SOLOv2 (Ours)&$\mathbf{40.2}$&$61.5$&$\mathbf{43.3}$&$17.4$&$\mathbf{43.3}$&$\mathbf{58.5}$ \\ 
    &RS-Mask R-CNN (Ours)&$40.6$&$62.8$&$43.9$&$22.8$&$43.6$&$52.8$ \\
    &RS-Mask R-CNN+ (Ours)&$\mathbf{42.0}$&$\mathbf{64.8}$&$\mathbf{45.6}$&$\mathbf{24.2}$&$\mathbf{45.1}$&$54.6$ \\ \hline
    \hhline{========}
    \multirow{5}{*}{\rotatebox{90}{w DCN}}&Mask-scoring R-CNN \cite{maskscoring}&$39.6$&$60.7$&$43.1$&$18.8$&$41.5$&$56.2$\\
    &BlendMask \cite{blendmask}&$41.3$&$63.1$&$44.6$&$22.7$&$44.1$&$54.5$\\
    &SOLOv2 \cite{solov2}&$41.7$&$63.2$&$45.1$&$18.0$&$45.0$&$\mathbf{61.6}$\\ \cline{2-8}
    &RS-Mask R-CNN+ (Ours)&$\mathbf{43.9}$&$\mathbf{67.1}$&$\mathbf{47.6}$&$\mathbf{25.6}$&$\mathbf{47.0}$&$57.8$\\ \cline{2-8}
    &RS-Mask R-CNN+* (Ours)&$44.8$&$68.4$&$48.6$&$27.1$&$47.9$&$58.3$\\ 
    %\cline{2-8} &RS-Mask R-CNN+$^*$ (Ours)&$45.3$&$68.6$&$49.4$&$28.5$&$48.4$&$57.4$\\ 
    %\cline{2-8}&RS-Mask R-CNN+ (MS Test)&$\mathbf{?}$&$?$&$\mathbf{?}$&$?$&$\mathbf{?}$&$\mathbf{?}$ \\ 
    \hline
    \end{tabular}
\end{table}

\subsection{Ablation Experiments}
\label{subsec:Ablation}
\textbf{Contribution of the components:} Replacing Focal Loss by RS Loss improves the performance significantly ($1$ AP - Table \ref{tbl:ablation}). Score-based weighting has a minor contribution and value-based task balancing simplifies tuning. 

\begin{table}
 \centering
\setlength{\tabcolsep}{0.25em}
\footnotesize
\caption{Contribution of the components of RS Loss on ATSS. \label{tbl:ablation}} 
\begin{tabular}{|c|c|c|c||c|c|}
\hline
Architecture&RS Loss&score-based w.&task bal.&$\mathrm{AP}$&H\# $\downarrow$\\ \hline
& & & &$38.7$&3\\ \cline{2-6}
ATSS+ResNet50&\checkmark& & &$39.7$&2\\ \cline{2-6}
w.o. aux. head&\checkmark&\checkmark& &$39.8$&2\\  \cline{2-6}
&\checkmark&\checkmark&\checkmark&$39.9$&1\\ \hline
\end{tabular}
\end{table}

\begin{table}
    \centering
\setlength{\tabcolsep}{0.35em}
    \footnotesize
    \caption{Ablation with different degrees of imbalance on different datasets and samplers. Number of negatives (neg) corresponding to a single positive (pos) averaged over the iterations of the first epoch is presented.
    %For each pos. instance (ins.), e.g. anchor, random (rand.) sampler aims to sample 1 neg. instance in RPN and 3 neg. instance in RCNN. With no sampler (none), there is no desired instance rate, i.e. not available - N/A. Note that when rand. sampler cannot find enough pos., the batch is padded with neg.; accordingly, actual instance rate presents number of neg. instances for each pos. instance after padding. Also, since we use class-wise binary classifiers on RCNN; each neg. instance has $O$ neg. tasks, and each pos. instance has $1$ pos. and $O-1$ neg. tasks where $O$ is the number of dataset classes (see Sec. 3.1 of Tan et al. [C] for details). Using this, actual class rate (underlined) presents the actual imbalance rate in the input of RS Loss as number of neg. tasks for a pos. task. 
    Quantitatively, pos:neg ratio varies between 1:7 to 1:10470. RS Loss successfully trains different degrees of imbalance without tuning (Tables \ref{tab:TwoStage} and \ref{tbl:LVIS}). Details: Appendix \ref{sec:Experiments_supp}}
    \label{tab:SamplingEffect}
    \begin{tabular}{|c|c|c|c|c|c|c|c|} \hline
    \multirow{2}{*}{Dataset}&\multicolumn{2}{|c|}{Sampler}&\multicolumn{2}{|c|}{Desired Neg \#}&\multicolumn{2}{|c|}{Actual Neg \#}&\multirow{2}{*}{$\mathrm{AP}$}\\ \cline{2-7}
    &RPN&R-CNN&RPN&R-CNN&RPN&R-CNN&\\ \hline
    \multirow{3}{*}{COCO}&Random&Random&1&3&7&702& 38.5\\ 
    &None&Random&1&N/A&6676&702&39.3\\
    &None&None&N/A&N/A&6676&1142&39.6 \\     \hline
    LVIS&None&None&N/A&N/A&3487&10470&25.2 \\     \hline
    \end{tabular}
\end{table}

\comment{
\begin{table}
    \centering
    \footnotesize
    \caption{RS Loss is robust to class imbalance. It successfully trains Faster R-CNN with both relatively balanced (``Random'' sampling) and severely imbalanced (``None'' in the table) data. Numbers in parentheses show positive to negative ratio of sampled examples.}
    \label{tab:SamplingEffect}
    \begin{tabular}{|c|c||c|c|c|} \hline
    RPN&R-CNN&AP&$\mathrm{AP_{50}}$&$\mathrm{AP_{75}}$\\ \hline 
    Random (1:1)&Random (1:3)&$38.5$&$58.5$&$41.5$ \\ 
    None&Random (1:3)&$39.3$&$\mathbf{59.6}$&$42.3$\\
    None&None&$\mathbf{39.6}$&$59.5$&$\mathbf{43.0}$ \\ \hline
    \end{tabular}
\end{table}
}

\textbf{Robustness to imbalance:} Without tuning, RS Loss can train models with very different imbalance levels successfully (Table \ref{tab:SamplingEffect}): Our RS Loss (i) yields $38.5$ AP on COCO with the standard random samplers (i.e. data is relatively balanced especially for RPN), (ii) utilizes more data when the samplers are removed, resulting in $\smallsim1$ AP gain ($38.5$ to $39.6$ AP), and (iii) outperforms all counterparts on the long-tailed LVIS dataset (c.f. Table \ref{tbl:LVIS}), where the imbalance is extreme for R-CNN (pos:neg ratio is $1:10470$ - Table \ref{tab:SamplingEffect}). Appendix \ref{sec:Experiments_supp} presents detailed discussion. 

% Moreover, while we train models with different distributions, unlike score-based losses (i.e. Focal Loss \cite{FocalLoss}, QFL \cite{GFL}), we do not tune the bias terms in the last layer of the classification head to prevent destabilization of the training due to large loss value originating from negatives, which is shown to be important for score-based loss functions \cite{SamplingHeuristics}.

\textbf{Contribution of the sorting error:} To see the contribution of our additional sorting error, during training, we track Spearman's ranking correlation coefficient ($\rho$) between IoUs and classification scores, as an indicator of the sorting quality, with and without our additional sorting error (see Eq. \ref{eq:RSLoss}-\ref{eq:RSTarget}). As hypothesized, using sorting error improves sorting quality, $\rho$, averaged over all/last 100 iterations, from $0.38/0.42$ to $0.42/0.47$ for RS-R-CNN.

\textbf{Effect on Efficiency:} On average, one training iteration of RS Loss takes around $1.5 \times$ longer than score-based losses. See Appendix \ref{sec:Experiments_supp} for more discussion on the effect of RS Loss on training and inference time.

%% file: sections/7.Conclusion.tex
\section{Conclusion}
In this paper, we proposed RS Loss as a ranking-based loss function to train object detectors and instance segmentation methods. Unlike existing ranking-based losses, which aim to rank positives above negatives, our RS Loss also sorts positives wrt. their localisation qualities, which is consistent with NMS and the performance measure, AP. With RS Loss, we  employed a simple, loss-value-based, tuning-free heuristic to balance all heads in the visual detectors. As a result, we showed on seven diverse visual detectors that RS Loss both consistently improves performance and significantly simplifies the training pipeline.
%We note that while our loss design presents a strong baseline wrt. performance, it is only validated on the common SOTA baseline ATSS, and the performance of other detectors may be improved further by a grid search or better task-balancing methods. Also, similar to other ranking-based losses \cite{APLoss}, the time and space complexity is quadratic since of RS Loss considers the ranking between each pair, and thus, training takes $\sim 1.5 \times$ longer on average compared to standard training. Nevertheless, we demonstrate on several detectors that one can achieve better performance than its score based alternatives using RS Loss only by tuning the learning rate and significantly simplifying training.

%% file: sections/8.Appendix.tex
%\appendix
%\addcontentsline{toc}{section}{Appendices}
%\maketitle
\section*{APPENDICES}

%\begin{abstract}
%This document provides details for and supporting the material in the paper. 
%\end{abstract}
\renewcommand{\thefigure}{A.\arabic{figure}}
\renewcommand{\thetable}{A.\arabic{table}}
\renewcommand{\theequation}{A.\arabic{equation}}
%\pagebreak
\renewcommand{\thesection}{A}
\input{appendix/1.DetailsofRSLoss}
\renewcommand{\thesection}{B}
\input{appendix/2.Analysis}
\renewcommand{\thesection}{C}
\input{appendix/3.MoreExperiments}
%\bibliography{detectionbibliography_appendix}

%% file: appendix/1.DetailsofRSLoss.tex
\section{Details of RS Loss}
\label{sec:RSLossDetails}
In this section, we present the derivations of gradients and obtain the loss value and gradients of RS Loss on an example in order to provide more insight.
\subsection{Derivation of the Gradients}
The gradients of a ranking-based loss function can be determined as follows. Eq. \ref{eq:Gradients} in the paper states that 
\begin{align}
    \label{eq:Gradients1}
    \frac{\partial \mathcal{L}}{\partial s_i} = \frac{1}{Z} \left( \sum \limits_{j \in \mathcal{P} \cup \mathcal{N} } \Delta x_{ji} - \sum \limits_{j \in \mathcal{P} \cup \mathcal{N}} \Delta x_{ij} \right). 
\end{align}

\noindent Our identity update reformulation suggests replacing $\Delta x_{ij}$ by $L_{ij}$ which yields:
\begin{align}
    \label{eq:GradientsPrimaryTerms}
    \frac{\partial \mathcal{L}}{\partial s_i} = \frac{1}{Z} \left( \sum \limits_{j \in \mathcal{P} \cup \mathcal{N}} L_{ji} - \sum \limits_{j \in \mathcal{P} \cup \mathcal{N}} L_{ij} \right). 
\end{align}
We split both summations into two based on the labels of the examples, and express $\frac{\partial \mathcal{L}}{\partial s_i}$ using four terms:
\begin{align}
    \label{eq:GradientsPrimaryTermsFourTerms}
    \frac{\partial \mathcal{L}}{\partial s_i} = \frac{1}{Z} \left( \sum \limits_{j \in \mathcal{P}} L_{ji} + \sum \limits_{j \in \mathcal{N}} L_{ji} - \sum \limits_{j \in \mathcal{P}} L_{ij} - \sum \limits_{j \in \mathcal{N}} L_{ij} \right). 
\end{align}
Then simply by using the primary terms of RS Loss, defined in Eq. 9 in the paper as:
\begin{align}
        \label{eq:RSPrimaryTermDefinition1}
        L_{ij} = \begin{cases} \left(\ell_{\mathrm{R}}(i) - \ell^*_{\mathrm{R}}(i) \right) p_{R}(j|i), & \mathrm{for}\; i \in \mathcal{P}, j \in \mathcal{N} \\
        \left(\ell_{\mathrm{S}}(i) - \ell^*_{\mathrm{S}}(i) \right) p_{S}(j|i), & \mathrm{for}\;i \in \mathcal{P}, j \in \mathcal{P},\\
        0, & \mathrm{otherwise},
        \end{cases}
\end{align}
With the primary term definitions,  we obtain the gradients of RS Loss using Eq. \ref{eq:GradientsPrimaryTermsFourTerms}.

\noindent\textbf{Gradients for $i \in \mathcal{N}$}. 
For $i \in \mathcal{N}$, we can respectively express the four terms in Eq. \ref{eq:GradientsPrimaryTermsFourTerms} as follows:
\begin{itemize}
    \item $\sum \limits_{j \in \mathcal{P}} L_{ji}=\sum \limits_{j \in \mathcal{P}}\left(\ell_{\mathrm{R}}(j) - \ell^*_{\mathrm{R}}(j) \right) p_{R}(i|j)$,
    \item $\sum \limits_{j \in \mathcal{N}} L_{ji}=0$ (no negative-to-negative error is defined for RS Loss -- see Eq. \ref{eq:RSPrimaryTermDefinition1}),
    \item $\sum \limits_{j \in \mathcal{P}} L_{ij}=0$ (no error when $j \in \mathcal{P}$ and $i \in \mathcal{N}$ for $L_{ij}$ -- see Eq. \ref{eq:RSPrimaryTermDefinition1}),
    \item $\sum \limits_{j \in \mathcal{N}} L_{ij}=0$ (no negative-to-negative error is defined for RS Loss -- see Eq. \ref{eq:RSPrimaryTermDefinition1}),
\end{itemize}
which, then, can be expressed as (by also replacing $Z=|\mathcal{P}|$ following the definition of RS Loss):
    \begin{align}
    \label{eq:RSNegGradients1}
   \frac{\partial \mathcal{L}_{\mathrm{RS}}}{\partial s_i} & = \frac{1}{|\mathcal{P}|} \left( \sum \limits_{j \in \mathcal{P}} L_{ji} + \cancelto{0}{\sum \limits_{j \in \mathcal{N}} L_{ji}} - \cancelto{0}{\sum \limits_{j \in \mathcal{P}} L_{ij}} - \cancelto{0}{\sum \limits_{j \in \mathcal{N}} L_{ij}} \right), \\ 
    & =\frac{1}{|\mathcal{P}|} \sum \limits_{j \in \mathcal{P}}\left(\ell_{\mathrm{R}}(j) - \cancelto{0}{\ell^*_{\mathrm{R}}(j)} \right) p_{R}(i|j), \\
    & =\frac{1}{|\mathcal{P}|}\sum \limits_{j \in \mathcal{P}} \ell_{\mathrm{R}}(j)p_{R}(i|j),
    \end{align}
concluding the derivation of the gradients if $i \in \mathcal{N}$. 

\noindent\textbf{Gradients for $i \in \mathcal{P}$}. 
We follow the same methodology for $i \in \mathcal{P}$ and express the same four terms as follows:
\begin{itemize}
    \item $\sum \limits_{j \in \mathcal{P}} L_{ji}=\sum \limits_{j \in \mathcal{P}} \left(\ell_{\mathrm{S}}(j) - \ell^*_{\mathrm{S}}(j) \right) p_{S}(i|j)$,
    \item $\sum \limits_{j \in \mathcal{N}} L_{ji}=0$ (no error when $j \in \mathcal{N}$ and $i \in \mathcal{P}$ for $L_{ji}$ -- see Eq. \ref{eq:RSPrimaryTermDefinition1}),
    \item $\sum \limits_{j \in \mathcal{P}} L_{ij}$ reduces to $\ell_{\mathrm{S}}(i) - \ell^*_{\mathrm{S}}(i)$ simply by rearranging the terms and $\sum \limits_{j \in \mathcal{P}} p_{S}(j|i)=1$ since $p_{S}(j|i)$ is a pmf:
    \begin{align}
        \sum \limits_{j \in \mathcal{P}} L_{ij}&=\sum \limits_{j \in \mathcal{P}} \left(\ell_{\mathrm{S}}(i) - \ell^*_{\mathrm{S}}(i) \right) p_{S}(j|i), \\
        &=\left(\ell_{\mathrm{S}}(i) - \ell^*_{\mathrm{S}}(i) \right) \cancelto{1}{\sum \limits_{j \in \mathcal{P}} p_{S}(j|i)}, \\
        &= \ell_{\mathrm{S}}(i) - \ell^*_{\mathrm{S}}(i).
    \end{align}
    \item Similarly, $\sum \limits_{j \in \mathcal{N}} L_{ij}$ reduces to $\ell_{\mathrm{R}}(i) - \ell^*_{\mathrm{R}}(i)$:
    \begin{align}
        \sum \limits_{j \in \mathcal{N}} L_{ij}&=\sum \limits_{j \in \mathcal{N}} \left(\ell_{\mathrm{R}}(i) - \ell^*_{\mathrm{R}}(i) \right) p_{R}(j|i) \\
        &=\left(\ell_{\mathrm{R}}(i) - \ell^*_{\mathrm{R}}(i) \right) \cancelto{1}{\sum \limits_{j \in \mathcal{N}} p_{R}(j|i)} \\
        &= \ell_{\mathrm{R}}(i) - \ell^*_{\mathrm{R}}(i).
    \end{align}
\end{itemize}
Combining these four cases together, we have the following gradient for $i \in \mathcal{P}$:
\begin{align}
\scriptsize
\label{eq:PositiveGrad}
 \frac{\partial \mathcal{L}_{\mathrm{RS}}}{\partial s_i} = &\frac{1}{|\mathcal{P}|} \left( \sum \limits_{j \in \mathcal{P}} \left(\ell_{\mathrm{S}}(j) - \ell^*_{\mathrm{S}}(j) \right) p_{S}(i|j) \right. \\
&\left. - \left( \ell_{\mathrm{S}}(i) - \ell^*_{\mathrm{S}}(i) \right)- \left( \ell_{\mathrm{R}}(i) - \ell^*_{\mathrm{R}}(i) \right) \right).
\end{align}
Finally, for clarity, we rearrange the terms also by using $\ell^*_{\mathrm{RS}}(i) - \ell_{\mathrm{RS}}(i)=-(\ell_{\mathrm{S}}(i)-\ell^*_{\mathrm{S}}(i))- (\ell_{\mathrm{R}}(i)-\ell^*_{\mathrm{R}}(i))$:
\begin{align}
\label{eq:PositiveGradFinal}
\frac{1}{|\mathcal{P}|} \left( \ell^*_{\mathrm{RS}}(i) - \ell_{\mathrm{RS}}(i) + \sum \limits_{j \in \mathcal{P}} \left(\ell_{\mathrm{S}}(j) - \ell^*_{\mathrm{S}}(j) \right) p_{S}(i|j) \right),
\end{align}
concluding the derivation of the gradients when $i \in \mathcal{P}$.

\subsection{More Insight on RS Loss Computation and Gradients on an Example}
\begin{figure}
    \centerline{
        \includegraphics[width=0.48\textwidth]{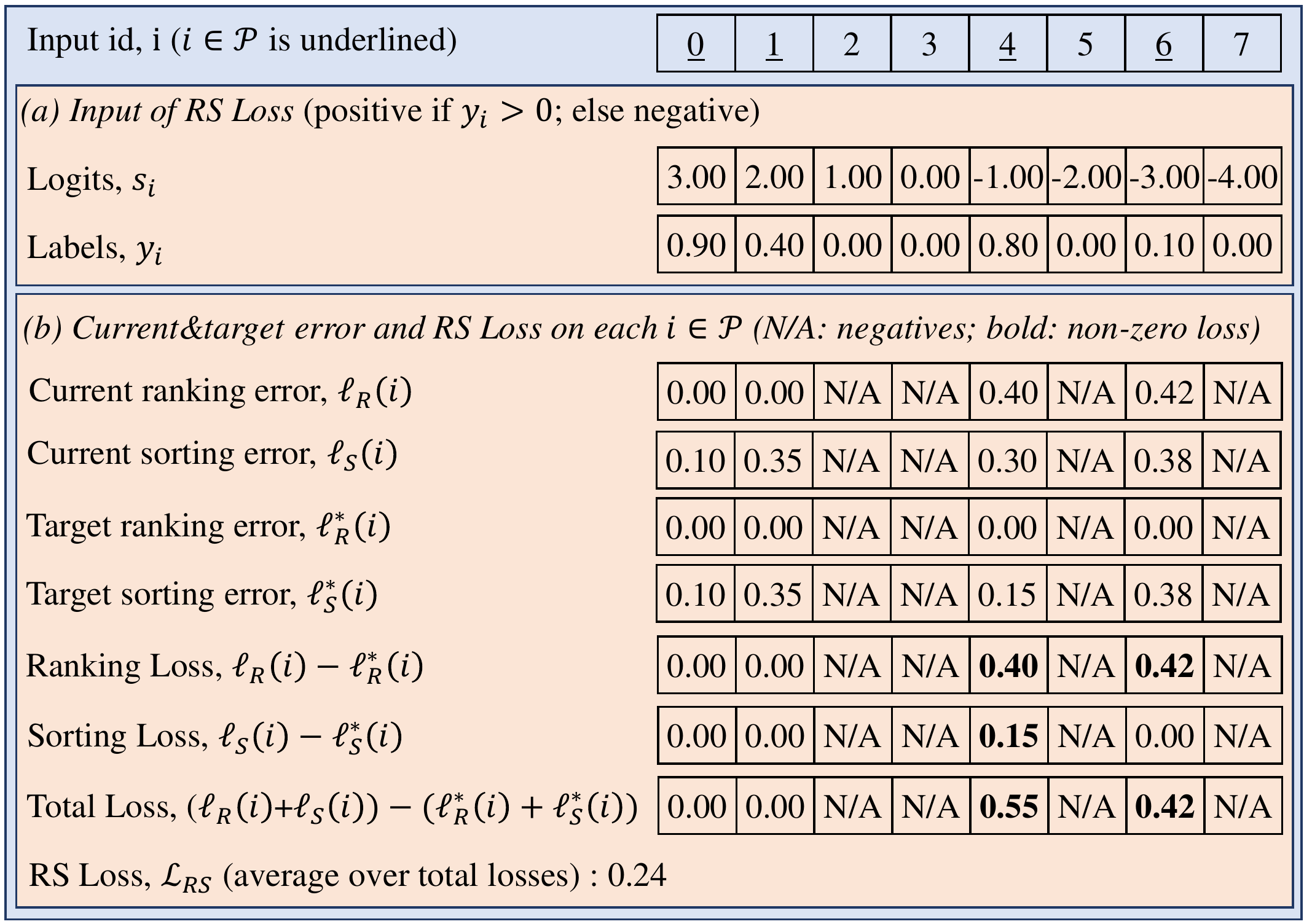}
    }
    \caption{An example illustrating the computation of RS Loss. (a) The inputs of RS Loss are logits and continuous labels (i.e. IoU). (b) Thanks to the ``Identity Update'' (Section 3.2 in the paper), the loss computed on each example  considers its  target error, hence, it is interpretable. E.g. $i=0, 1, 6$ have positive current sorting errors, but already sorted properly among examples with larger logits, which can be misleading when loss value ignores the target error. Since RS Loss is computed only over positives, N/A is assigned for negatives.  
    %(c) The losses are then distributed over examples causing sorting and ranking errors via pmfs $p_S(j|i)$ and $p_R(j|i)$ to determine pairwise errors, i.e. primary terms, and hence the gradients following identity update (see  Eq. \ref{eq:RSNegGradients1} and \ref{eq:RSPosGradients}; and Supp.Mat. for details.). The gradients of RS Loss considers soft labels and balanced training:  (i) Thanks to its sorting objective, RS Loss assigns a gradient to suppress a positive example when it is not ranked among positives accordingly wrt its soft label, and (ii) total gradient magnitudes of positives and negatives are equal, and larger gradients are assigned to harder negatives (e.g. for $i=7 \in \mathcal{N}, \frac{\partial \mathcal{L}}{\partial s_7}=0$).
    \label{fig:RSLoss_def}
} 
\end{figure}
\begin{figure}
    \centerline{
        \includegraphics[width=0.48\textwidth]{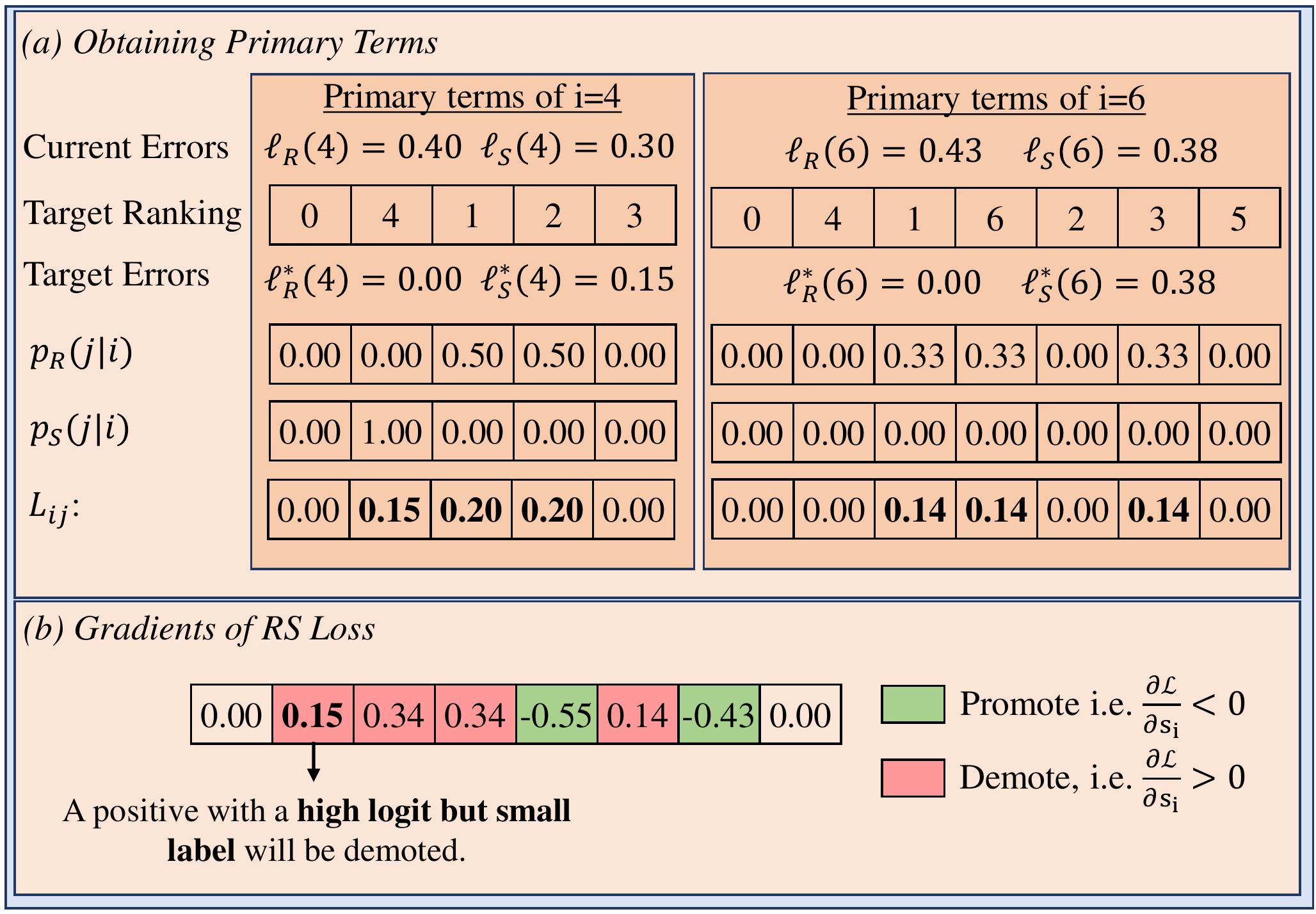}
    }
    \caption{An example illustrating the computation of primary terms in RS Loss. (a) The computation of primary terms. We only show the computation for positives $i=4$ and $i=6$ since for $i=0$ and $i=1$ the total loss is $0$ (see Fig. \ref{fig:RSLoss_def}(b)); and RS Loss does not compute error on negatives by definition (i.e. discretizes the space only on positives). To compute primary terms, $L_{ij}$, one needs current errors, target errors and pmfs  for both ranking and sorting, which are included in the figure respectively. In order to compute the target errors on a positive $i \in \mathcal{P}$, the examples are first thresholded from $s_i$ and the ones with larger (i.e. $s_j \geq s_i$) logits are obtained. Then, target rankings are identified using continuous labels. The ranking and sorting errors computed for the target ranking determines target errors, $\ell_R^*(i)$ and $\ell_S^*(i)$. The ranking and sorting losses, $\ell_R(i)-\ell_R^*(i)$ and $\ell_S(i)-\ell_S^*(i)$ respectively, are then distributed over examples causing these losses uniformly via pmfs $p_S(j|i)$ and $p_R(j|i)$ to determine pairwise errors, i.e. primary terms. (b) The gradients are obtained simply by using primary terms as the update in Eq. \ref{eq:Gradients} following identity update yielding  Eq. \ref{eq:RSNegGradients} and \ref{eq:PositiveGradFinal} for negatives and positives respectively. Thanks to the novel sorting objective, RS Loss can assign a gradient to suppress a positive example when it is not ranked among positives accordingly wrt its continuous label (e.g. $i=1$).
    \label{fig:RSLoss_comp}
} 
\end{figure}

In Fig.  \ref{fig:RSLoss_def}, we illustrate the input and the  computation of RS Loss. We emphasize that our Identity Update provides interpretable loss values when the target value is non-zero (Fig. \ref{fig:RSLoss_def}(b)). Previous work \cite{APLoss, aLRPLoss} fail to provide interpretable loss values.

\textbf{Computation of the Loss.} To compute the loss following the three-step algorithm in the paper, the first and the third steps are trivial (Fig. 2 in the paper), thus, here we present in Fig. \ref{fig:RSLoss_comp} (a) how the primary terms ($L_{ij}$) are computed for our example (Fig. \ref{fig:RSLoss_def}) following Eq. \ref{eq:RSPrimaryTermDefinition1}.

\textbf{Optimization of the Loss.} Fig. \ref{fig:RSLoss_comp}(b) presents and discusses the gradients obtained using Eq. \ref{eq:RSNegGradients1}, \ref{eq:PositiveGradFinal}. 

%% file: appendix/2.Analysis.tex
\section{Analyses}
\label{sec:Analysis}
Section \ref{subsec:DesignChoices} presents our experiments to validate our design choices and Section \ref{subsec:aLRPLoss} discusses the drawbacks of aLRP Loss, and how we fix them.

\subsection{Analysis to Determine Design Choices in Localisation Loss}
\label{subsec:DesignChoices}
In this section, we provide our analysis on ATSS \cite{ATSS} to determine our design choices for localisation. First, as a baseline, we train ATSS network with the following loss function:
\begin{align}
    \label{eq:LossATSSRS1}
    \mathcal{L}_{RS-ATSS}=\mathcal{L}_{RS}+ \lambda_{box} \mathcal{L}_{box},
\end{align}
where $\mathcal{L}_{RS}$ is our Rank \& Sort Loss, $\lambda_{box}$ is the task-level balancing coefficient and $\mathcal{L}_{box}$ is the box regression loss.

First, we investigate two tuning-free heuristics to determine  $\lambda_{box}$ every iteration: (i) value-based: $\lambda_{box} = \mathcal{L}_{RS}/\mathcal{L}_{box}$, and (ii) magnitude-based: $\lambda_{box}= \abs{ \frac{{\partial \mathcal{L}}_{\mathrm{RS}}}{{\partial \mathbf{\hat{s}}}}} / \abs{ \frac{{\partial \mathcal{L}}_{\mathrm{box}}}{{\partial \mathbf{b}}}}$ where $\abs{\cdot}$ is L1 norm, $\mathbf{\hat{b}}$ and $\mathbf{s}$ are box regression and classification head outputs respectively. Table \ref{tab:localisation} presents that value-based task balancing performs similar to tuning $\lambda_{box}$ ($\sim 0$ AP on average).

Secondly, we delve into $\mathcal{L}_{box}$, which is defined as the weighted average of the individual losses of examples:
\begin{align}
\label{eq:regressionloss}
    \mathcal{L}_{box}= \sum \limits_{i \in \mathcal{P}} \frac{w^i}{\sum \limits_{j \in \mathcal{P}} w^j} \mathcal{L}_{GIoU} (\hat{b}_i, b_i),
\end{align}
where $\mathcal{L}_{GIoU}(\hat{b}_i, b_i)$ is the GIoU Loss \cite{GIoULoss}, and $w^i$ is the \textit{instance-level importance weight}. Unlike \textit{no prioritization} (i.e. $w^i=1$ for $i \in \mathcal{P}$), recently, a diverse set of heuristics assigns different importances over $i \in \mathcal{P}$: \textit{centerness-based} importance \cite{FCOS,ATSS} aims to focus on the proposals (i.e. point or anchor) closer to the center of $b_i$, \textit{score-based} heuristic \cite{GFL} uses the maximum of confidence scores of a prediction as $w^i$,  \textit{IoU-based} approach \cite{paa} increases the losses of the predictions that are already better localized by $w^i=\mathrm{IoU}(\hat{b}_i, b_i)$, and finally \textit{ranking-based} weighting \cite{aLRPLoss} uses  $w^i=\frac{1}{|\mathcal{P}|} \left( {\sum \limits_{k \in \mathcal{P}}}  \frac{\mathrm{H}(x_{ki})}{\mathrm{rank}(k)}\right)$, where $\mathrm{H}(\cdot)$ can be smoothed by an additional hyper-parameter ($\delta_{loc}$). Note that these instance-level weighting methods perform similarly (largest gap is $0.2$ AP -- Table \ref{tab:localisation}) and we prefer score-based weighting with RS Loss.

% \textbf{Observations and Our Design Choices}: In our experiments on ATSS trained with RS Loss (Table \ref{tab:localisation}), we observed that: (i) value-based task balancing performs similar to tuning $\lambda_{box}$ ($\sim 0$ AP on average), (ii) instance-level weighting methods also perform similarly (largest gap is $0.2$ AP). Thus, we use value-based task balancing and score-based instance weighting, which are both hyper-parameter-free and easily applicable to all networks. With these design choices, Eq. \ref{eq:LossATSSRS1} has only $1$ hyper-parameter (i.e. $\delta_{RS}$ in $\mathrm{H}(\cdot)$, set to $0.50$, to smooth the unit-step function).

\comment{
Table \ref{tab:localisation} presents the experiments that we conducted on ATSS \cite{ATSS} using our RS Loss to make  design choices regarding localisation loss. Since different approaches yield similar performances, we prefer
\begin{itemize}
    \item score-based weighting as the instance-level importance weight, and
    \item value-based balancing as the task-level balance coefficient,
\end{itemize}
which are both hyperparameter-free.
}
\begin{table}
    \centering
    \setlength{\tabcolsep}{0.25em}
    \footnotesize
    \caption{Comparison of instance- and task-level weighting methods on RS-ATSS. Instance-level importance weighting methods yield similar performance and value-based SB achieves similar performance with constant weighting. Thus, we use score-based weighting and value-based SB with RS Loss (underlined\&bold), which are both tuning-free. }
    \label{tab:localisation}
    \begin{tabular}{|c|c|c|c|c|c|} \hline
    \multirow{3}{*}{\begin{minipage}[c][1.3cm][c]{0.175\textwidth}\centering Instance-level importance weight ($w^i$)
 \end{minipage}}&\multicolumn{5}{|c|}{Task-level balancing coefficient ($\lambda_r$)} \\ \cline{2-6}
    &\multicolumn{3}{|c|}{Constant weighting}&\multicolumn{2}{|c|}{Self-balance (SB)} \\ \cline{2-6}
    &$1$&$2$&$3$&value&magnitude\\ \hline \hline
    No prioritization&$38.9$&$39.7$&$39.7$&$39.7$&$39.4$\\ \hline
    Centerness-based \cite{FCOS}&$38.8$&$39.8$&$39.6$&$39.6$&$39.5$\\ \hline
    Score-based \cite{GFL}&$39.1$&$39.8$&$39.7$& \underline{$\mathbf{39.9}$}&$39.7$\\ \hline
    IoU-based \cite{paa}&$39.0$&$39.7$&$39.8$&$39.7$&$39.6$\\ \hline
    Ranking-based \cite{aLRPLoss}&$39.1$&$\mathbf{39.9}$&$39.6$&$\mathbf{39.9}$&$39.6$\\ \hline
    \end{tabular}
\end{table}

\subsection{A Comparative Analysis with  aLRP Loss}
\label{subsec:aLRPLoss}
In this section, we list our observations on aLRP Loss \cite{aLRPLoss} based on our comparative analysis with RS Loss: 

\begin{table}
    \centering
    \caption{Due to epoch-based self-balance and competition of tasks for the limited range, aLRP Loss performs significantly worse in the first epoch. When the model is trained longer using heavy training (i.e. 100 epochs, SSD-style augmentation \cite{SSD}), the default configuration of aLRP Loss, the performance gap relatively decreases at the end of training, however, the gap is still significant ($\sim 2$ AP) for the standard training (i.e. 12 epochs, no SSD-style augmentation \cite{SSD}). All experiments are conducted on Faster R-CNN.}
    \label{tab:aLRPLossDrawbacks}
    \setlength{\tabcolsep}{0.25em}
    \begin{tabular}{|c|c|c|c|c|} \hline
    \multirow{2}{*}{\begin{minipage}[c][0.7cm][c]{0.10\textwidth}\centering Loss Function
 \end{minipage}}&\multicolumn{2}{|c|}{Heavy Training} &\multicolumn{2}{|c|}{Standard Training} \\ \cline{2-5}
     &Epoch 1&Epoch 100&Epoch 1&Epoch 12\\ \hline
    %CE\&L1 Losses &$15.4$&$37.8$&$20.2$&$37.4$ \\ \hline
    aLRP Loss \cite{aLRPLoss}&$9.4$&$40.7$&$14.4$&$37.4$\\ \hline
    RS Loss &$\mathbf{17.7}$&$\mathbf{41.2}$&$\mathbf{22.0}$& $\mathbf{39.6}$\\ \hline
    \end{tabular}
\end{table}

\textbf{Observation 1: Tasks competing with each other within the bounded range of aLRP Loss degrades performance especially when the models are trained 12 epochs following the common training schedule.} 

To illustrate this, we train Faster R-CNN \cite{FasterRCNN} with aLRP Loss and our RS Loss using two different settings: 
\begin{itemize}
    \item ``Standard Training'', which refers to the common training (e.g. \cite{ATSS,FasterRCNN,FCOS}): The network is trained by a batch size of 16 images with resolution $1333 \times 800$ without any augmentation except the standard horizontal flipping. We use 4 GPUs, so each GPU has 4 images during training. We tune the learning rate of aLRP Loss as $0.009$ and for our RS Loss we set it to $0.012$. Consistent with the training image size, the test image size is $1333 \times 800$. 
    
    \item ``Heavy Training'', which refers to the standard training design of aLRP Loss (and also AP Loss): The network is trained by a batch size of 32 images with resolution $512 \times 512$ on 4 GPUs (i.e. 8 images/GPU) using SSD-syle augmentation \cite{SSD} for 100 epochs. We use the initial learning rate of $0.012$ for aLRP Loss as validated in the original paper, and for our RS Loss, we simply use linear scheduling hypothesis and set it to $0.024$ without further validation. Here, following aLRP Loss (and AP Loss), the test image size is $833 \times 500$.  
\end{itemize}

Table \ref{tab:aLRPLossDrawbacks} presents the results and we observe the following:
\begin{enumerate}
    \item \textit{For both ``heavy training'' and ``standard training'', aLRP Loss has significantly lower performance after the first epoch ($17.7$ AP vs. $9.2$ AP for heavy training and $22.0$ AP vs. $14.4$ AP) compared to RS Loss}: aLRP Loss has a bounded range between $0$ and $1$, which is dominated by the classification head especially in the beginning of the training, and hence, the box regression head is barely trained. To tackle that, Oksuz et al. \cite{aLRPLoss} dynamically promotes the loss of box regression head  using a self-balance weight, initialized to $50$ and updated based on loss values at the end of every epoch. However, we observed that this range pressure has an adverse effect on the performance especially at the beginning of the training, which could not be fully addressed by self-balance since in the first epoch the SB weight is not updated.
    
    \item \textit{While the gap between RS Loss and aLRP Loss is $0.5$ AP for  ``heavy training'', it is $2.2$ AP for ``standard training''.} After the SB weight of aLRP Loss is updated, the gap can be reduced when the models are trained for longer epochs. However, the final gap is still large ($\sim 2$ AP) for ``standard training'' with 12 epochs since unlike aLRP Loss, our RS Loss (i) does not have a single bounded range for which multiple tasks compete, and (ii) uses an iteration-based self-balance instead of epoch-based. 
\end{enumerate}

\textbf{Observation 2: The target of aLRP Loss does not have an intuitive interpretation.} 

Self-balance (or range pressure --  see Observation 1) is not the single reason why RS Loss performs better than aLRP Loss in both scheduling methods in Table \ref{tab:aLRPLossDrawbacks}. aLRP Loss uses the following target error for a positive example $i$:
\begin{align}
  \ell_{aLRP}^*(i)=\frac{\mathcal{E}_{loc}(i)}{\mathrm{rank}(i)},
\end{align}
where
\begin{align}
    \mathcal{E}_{loc}(i)=\frac{1-\mathrm{IoU}(\Hat{b}_i, b_i)}{1-\tau},\label{eqn:tau}
\end{align}
and $\tau$ is the positive-negative assignment threshold. However, unlike the target of RS Loss for specifying the error at the target ranking where positives are sorted wrt their IoUs (see Fig. \ref{fig:RSLoss_comp}), the target of aLRP Loss does not have an intuitive interpretation.

\textbf{Observation 3: Setting $\tau$ in Eq. \ref{eqn:tau} to the value of the positive-negative (anchor IoU) assignment threshold creates ambiguity (e.g. anchor-free detectors do not have such a  threshold).} 

We identify three obvious reasons: (i) Anchor-free methods do not use IoU to assign positives and negatives, (ii) recent SOTA anchor-based methods, such as ATSS \cite{ATSS} and PAA \cite{paa}, do not have a sharp threshold to assign positives and negatives, but instead they use adaptive thresholds to determine positives and negatives during training, and furthermore (iii) anchor-based detectors split anchors as positives and negatives; however, the loss is computed on the predictions which may have less IoU with ground truth than $0.50$. Note that our RS Loss directly uses IoUs as the continuous labels without further modifying or thresholding them.

\textbf{Observation 4: Using an additional hyper-parameter ($\delta_{loc}$) for ranking-based weighting yields better performance for the common 12 epoch training.} 

As also discussed in Section \ref{subsec:DesignChoices}, \textit{ranking-based} importance weighting of the instances corresponds to:
\begin{align}
w^i=\frac{1}{|\mathcal{P}|} \left( {\sum \limits_{k \in \mathcal{P}}}  \frac{\mathrm{H}(x_{ki})}{\mathrm{rank}(k)}\right).    
\end{align}
aLRP Loss, by default, prefers not to smooth the nominator ($\mathrm{H}(x_{ki})$) but $\mathrm{rank}(k)$ is computed by the smoothed unit-step function. We label this setting as ``default'' and introduce an additional hyper-parameter $\delta_{loc}$ to further analyse ranking-based weighting. Note that the larger $\delta_{loc}$ is, the less effect the logits will have on $w^i$ (Eq. \ref{eq:regressionloss}). In Table \ref{tab:Delta_loc}, we compare these different settings on RS-ATSS trained for 12 epochs with our RS Loss, and observe that the default ranking-based weighting can be improved with different $\delta_{loc}$ values. However, for our RS Loss, we adopt score-based weighting owing to its tuning-free nature.

\begin{table}
    \centering
    \caption{Using an additional $\delta_{loc}$ to smooth the effect of ranking-based weighting can contribute to the performance.}
    \label{tab:Delta_loc}
    \begin{tabular}{|c||c|c|c|c|c|c|} \hline
    $\delta_{loc}$&$0.00$&$0.50$& $1.00$& $1.50$& $2.00$&Default\\ \hline
    AP &$39.3$&$39.4$&$39.8$&$\mathbf{39.9}$&$39.8$&$39.5$\\ \hline
    \end{tabular}
\end{table}

%% file: appendix/3.MoreExperiments.tex
\section{More Experiments on RS Loss}
\label{sec:Experiments_supp}
This section presents the experiments that are omitted from the paper due to space constraints.

\subsection{Effect of $\delta_{RS}$, the Single Hyper-parameter, for RS Loss.} 
Table \ref{tab:Delta} presents the effect of $\delta_{RS}$ on RS Loss using ATSS. We observe similar performance between $\delta_{RS}=0.40$ and $\delta_{RS}=0.75$. Also note that considering positive-to-positive errors in the sorting error, we set $\delta_{RS}$ different from AP Loss and aLRP Loss, both of which smooth the unit step function by using $\delta_{RS}=1.00$ as validated by Chen et al. \cite{APLoss}.

\begin{table}
    \centering
    \caption{We set $\delta_{RS}=0.50$, the only hyper-parameter of RS Loss, in all our experiments.}
    \label{tab:Delta}
    \begin{tabular}{|c||c|c|c|c|c|c|} \hline
    $\delta_{RS}$&$0.25$&$0.40$& $0.50$& $0.60$& $0.75$&$1.00$\\ \hline
    AP &$39.0$&$39.7$&$\mathbf{39.9}$&$39.7$&$39.8$&$39.4$\\ \hline
    \end{tabular}
\end{table}

\subsection{Training Cascade R-CNN \cite{CascadeRCNN} with RS Loss} 
Table \ref{tab:Cascaded} shows that using RS Loss to train Cascade R-CNN (RS-Cascade R-CNN) also improves baseline Cascade R-CNN by $1.0$ AP. We note that unlike the conventional training, we do not assign different loss weights over each R-CNN.

\begin{table}
    \centering
    \caption{RS Loss improves strong baseline Cascade R-CNN \cite{CascadeRCNN}.}
    \setlength{\tabcolsep}{0.4em}
    \label{tab:Cascaded}
    \begin{tabular}{|c||c|c|c|c|} \hline
    Method&AP $\uparrow$&$\mathrm{AP_{50}} \uparrow$&$\mathrm{AP_{75}} \uparrow$&oLRP $\downarrow$\\ \hline
    Cascade R-CNN&$40.3$&$58.6$&$44.0$&$67.0$\\ \hline
    RS Cascade R-CNN&$\textbf{41.3}$&$\textbf{58.9}$&$\textbf{44.7}$&$\textbf{66.6}$ \\ \hline
    \end{tabular}
\end{table}

\subsection{Hyper-parameters of R-CNN Variants in Table 1 of the Paper} 
A two-stage detector that uses random sampling and does not employ a method to adaptively set $\lambda_{t}^k$ has at least 7 hyper-parameters since (i) for random sampling, one needs to tune number of foreground examples and number of background examples to be sampled in both stages (4 hyper-parameters), and (ii) at least 3 $\lambda_{t}^k$s need to be tuned as the task-balancing coefficients in a loss with four components (Eq. 1 in the paper). As a result, except aLRP Loss and our RS Loss, all methods have at least 7 hyper-parameters. When the box regression losses of RPN and R-CNN are L1 Loss, GIoU Loss or AutoLoss, and the network has not an additional auxiliary head, 7 hyper-parameters aree sufficient (i.e. GIoU Loss \cite{GIoULoss}, Carafe FPN \cite{carafe} and AutoLoss-A \cite{autoloss}). Below, we list the methods with more than 7 hyper-parameters:

\begin{itemize}
    \item FPN \cite{FeaturePyramidNetwork} uses Smooth L1 in both stages, resulting in 2 more additional additional hyper-parameters ($\beta$) to be tuned for the cut-off from L2 Loss to L1 Loss for Smooth L1 Loss.
    \item IoU-Net \cite{IoUNet} also has Smooth L1 in both stages. Besides, there is an additional IoU prediction head trained also by Smooth L1, implying  $\lambda_{t}^k$ for IoU prediction head and $\beta$ for Smooth L1. In total, there are 7 hyper-parameters in the baseline model, and with these 4 hyper-parameters, IoU-Net includes 11 hyper-parameters.
    \item To train R-CNN, Libra R-CNN \cite{LibraRCNN} uses IoU-based sampler, which splits the negatives into IoU bins with an IoU interval width of $\kappa$, then also exploits random sampling. Besides it uses Balanced L1 Loss which adds 2 more hyper-parameters to Smooth L1 Loss (3 hyper-parameters in total). As a result, Libra R-CNN has 11 hyper-parameters in sampling and loss function in total.
    %\item KL Loss \cite{KLLoss} uses Smooth L1 for RPN and initializes the mean and variance of KL Loss. Hence with these 3 additional it has 10 hyper-parameters in total.
    \item Dynamic R-CNN \cite{dynamicrcnn} uses Smooth L1 for RPN and adds one more hyper-parameter to the Smooth L1, resulting in 3 additional hyperparameters. As a result, it has 10 hyper-parameters.
\end{itemize}

\subsection{Using Different Localisation Qualities as Continuous Labels to Supervise Instance Segmentation Methods} 
In order to provide more insight regarding the employment of continuous labels for the instance segmentation methods, we train YOLACT under four different settings: (i) without using continuous labels (c.f. ``Binary'' in Table \ref{tab:softlabeleffect}) (ii) using IoU, the bounding box quality, as the continuous label (iii) using Dice coefficient, the segmentation quality, as the continuous label and (iv) using the average of IoU and Dice coefficient as the continuous label. Table \ref{tab:softlabeleffect} suggests that all of these localisation qualities improve performance against ignoring them during training. Therefore, we use IoU as the continuous ground truth labels in all of our experiments with the exception of RS-SOLOv2, in which we used Dice coefficient, yielding similar performance to using IoU (Table \ref{tab:softlabeleffect}), since SOLOv2 does not have a box regression head.

\begin{table}
    \centering
    \caption{Analysis whether using continuous labels is useful for instance segmentation. We use IoU to supervise instance segmentation methods except SOLOv2, in which we use Dice coefficient since bounding boxes are not included in the output. Using Dice coefficient also provides similar performance with IoU. Binary refers to the conventional training (i.e. only ranking without sorting) without continuous labels.}
    \setlength{\tabcolsep}{0.3em}
    \label{tab:softlabeleffect}
    \begin{tabular}{|c||c|c|c||c|c|c|} \hline
\multirow{2}{*}{Label}&\multicolumn{3}{|c||}{Segmentation}&\multicolumn{3}{|c|}{Detection}\\ \cline{2-7}
    &AP&$\mathrm{AP_{50}}$&$\mathrm{AP_{75}}$&AP&$\mathrm{AP_{50}}$&$\mathrm{AP_{75}}$\\ \hline 
    Binary&$29.1$&$49.9$&$29.4$&$32.9$&$53.8$&$34.2$\\ \hline 
    IoU&$\mathbf{29.9}$&$\mathbf{50.5}$&$\mathbf{30.6}$&$\mathbf{33.8}$&$54.2$&$\mathbf{35.4}$\\ 
    Dice&$29.8$&$50.4$&$30.2$&$33.5$&$\mathbf{54.3}$&$35.1$\\ 
    (IoU+Dice)/2&$29.6$&$50.2$&$30.0$&$33.4$&$54.1$&$34.8$\\ \hline
    \end{tabular}
\end{table}

\subsection{Details of the Ablation Analysis on Different Degrees of Imbalance}
This section presents details on the discussion on robustness of RS Loss to imbalance (Section 6.3 in the paper). 

\textbf{Experimental Setup.} Using RS Loss on multi-stage visual detectors (e.g. Faster R-CNN or Mask R-CNN) involves two major changes in the training pipeline:
\begin{enumerate}
    \item The random samplers from both stages (i.e. from RPN and R-CNN) are removed.
    \item The $O+1$-way softmax classifier, where $O$ is the number of classes in the dataset, is replaced by $O$ binary (i.e. class-wise) sigmoid classifiers for the second stage of Faster R-CNN (i.e. R-CNN)\footnote{Note that RPN, which aims to determine ``objectness'', is already implemented by a single sigmoid classifier in mmdetection \cite{mmdetection}. Hence, no modification is required for the classifier of RPN.}.
\end{enumerate}

Note that in order to present the actual imbalance ratio between positives (pos) and negatives (neg), one needs to track the actual \textit{task} ratio resulting from the binary sigmoid classifiers. That is, with $O$ individual binary sigmoid classifiers, each positive instance (e.g. anchor, proposal/region-of-interest) yields $1$ pos and $O-1$ neg tasks, and each negative instance yields $O$ negative tasks (also refer to Section 3.1 of Tan et al. \cite{Eqv2} for details). To illustrate (Table \ref{tab:SamplingEffect1}), when we aim 1:3 pos:neg instance ratio for R-CNN by using a random sampler, as conventionally done, the actual \textit{instance pos:neg ratio} turns out to be 1:8 since the sampler pads the fixed batch size (i.e. in terms of proposals/regions-of-interest, which is 256 in this case) with negative instances when there is no enough positives. On the other hand, the actual \textit{task pos:neg ratio} is 1:702, implying that the pos:neg ratio of instances is not representative. As a result, we consider the actual task pos:neg ratio as the actual imbalance ratio.

\textbf{Robustness of RS Loss to Imbalance.} In order to show that RS Loss is robust to different degrees of imbalance without tuning, we trained (i) three Faster R-CNN \cite{FasterRCNN} on COCO dataset \cite{COCO} by gradually removing the random sampler from both stages and also (ii) one Mask R-CNN on LVIS dataset \cite{LVIS} as an extremely imbalanced case. Table \ref{tab:SamplingEffect1} presents pos:neg instance and task ratios averaged over the iterations during the first epoch\footnote{Note that since the anchors, fed to the first stage (i.e. RPN), are fixed in location, scale and aspect ratio during the training, the imbalance ratios in the first epoch also applies for all epochs for RPN; on the other hand, for R-CNN the number of negatives for each positive may increase in the latter epochs since the RPN will be able to classify and locate more objects.}:

\begin{itemize}
    \item When the random samplers are removed from both stages, the actual pos:neg task ratio increases. Specifically, due to the large number of anchors used for training RPN, actual pos:neg task ratio increases significantly for RPN (from 1:7 to 1:6676). As for R-CNN, this change is not as dramatic as RPN on COCO dataset after the sampler is removed (from 1:702 to 1:1142 -- compare ``Random'' and ``None'' for R-CNN in Table \ref{tab:SamplingEffect1}) since R-CNN is trained with top-1000 scoring region-of-interests (instead of all anchors in RPN) and COCO dataset has 80 classes. Note that RS Loss can train all three configurations (whether random sampling is removed or not) for COCO dataset successfully, and when more data is available (i.e. sampler is ``None''), the performance improves from $38.5$ to $39.6$.
    \item When we train Mask R-CNN using RS Loss on the long-tailed LVIS dataset without any samplers, we observed that unlike COCO dataset, R-CNN training is \textit{extremely imbalanced} (actual pos:neg task ratio is 1:10470) due to the large number of classes in LVIS dataset. Still, our RS Loss achieves SOTA performance despite this extreme imbalance (see also Table 5 in the paper). 
\end{itemize}

As a result, we conclude that RS Loss can easily be incorporated to train data with different levels of imbalance.

\begin{table*}
    \centering
    \caption{Ablation with different degrees of imbalance. Positive:negative (pos:neg) ratios averaged over the iterations of the first epoch of training with RS Loss on different datasets \& samplers. For each pos instance, e.g. anchor, random sampler aims to sample 1 neg instance in RPN and 3 neg instance in R-CNN. With no sampler (None), there is no desired pos:neg instance ratio, i.e. not available - N/A. Note that when random sampler cannot find enough pos, the batch is padded with neg; accordingly, actual pos:neg instance ratio is computed after this padding. Since we use binary sigmoid classifiers on R-CNN; each neg instance has $O$ neg tasks, and each pos instance has $1$ pos and $O-1$ neg tasks where $O$ is the number of dataset classes. Using this, actual pos:neg task ratio (underlined) presents the actual imbalance ratio in the input of RS Loss. Quantitatively, the actual pos:neg task ratio varies between 1:7 to 1:10470. Despite very different degrees of imbalance, our RS Loss outperforms all counterparts without sampling and tuning in both datasets.}
    \label{tab:SamplingEffect1}
    \setlength{\tabcolsep}{0.3em}
    \begin{tabular}{|c|c|c|c|c|c|c|c|c||c|} \hline
    \multirow{2}{*}{Dataset}&\multicolumn{2}{|c|}{Sampler}&\multicolumn{2}{|c|}{desired pos:neg instance ratio}&\multicolumn{2}{|c|}{actual pos:neg instance ratio}&\multicolumn{2}{|c|}{\underline{actual pos:neg task ratio}}&\multirow{2}{*}{$\mathrm{AP}$}\\ \cline{2-9}
    &RPN&R-CNN&RPN&R-CNN&RPN&R-CNN&\underline{RPN}&\underline{R-CNN}&\\ \hline
    \multirow{3}{*}{COCO}&Random&Random&1:1&1:3&1:7&1:8&\underline{1:7}&\underline{1:702}& 38.5\\ 
    &None&Random&1:1&N/A&1:6676&1:8&\underline{1:6676}&\underline{1:702}&39.3\\
    &None&None&N/A&N/A&1:6676&1:13&\underline{1:6676}&\underline{1:1142}&39.6 \\     \hline
    LVIS&None&None&N/A&N/A&1:3487&1:12&\underline{1:3487}&\underline{1:10470}&25.2 \\     \hline
    \end{tabular}
\end{table*}

\textbf{Are Score-based Loss Functions Robust to Imbalance Without Tuning?} Here, we investigate how Cross-entropy Loss and Focal Loss behave when samplers are removed without any tuning. 

\textit{Cross-entropy Loss.} As a fair baseline for our RS Loss, we use Faster R-CNN with GIoU Loss and only remove random sampling gradually similar to how we did for RS Loss. Table \ref{tab:SamplingEffect1Score} shows that, as opposed to our RS Loss, the performance significantly drops once the samplers are removed for Cross-entropy Loss, and hence Cross-entropy Loss cannot be directly employed to train different levels of imbalance unlike our RS Loss.

\textit{Focal Loss.} There are many design choices that one needs to tune in order replace the standard Cross-entropy Loss by Focal Loss. Here, instead of tuning each of these extensively, we use a commonly used setting in one-stage detectors \cite{ATSS,GFL,FocalLoss} to train RPN and R-CNN. In particular, we use individual class-wise binary sigmoid classifier (as we also did for RS Loss), set the learning rate to $0.01$, the weight of GIoU Loss to $2$ and the bias terms in the last layer of the classification head\footnote{Note that we also do not tune this bias term for our RS Loss and use the default setting for all the detectors that we train.} such that the confidence scores of the positives are $0.01$ to prevent destabilization of the training due to large loss value originating from negatives. However, we observed that Focal Loss is not able to perform as good as Cross-entropy Loss and RS Loss with this configuration (Table \ref{tab:SamplingEffect1}). Hence, as a generalisation of Cross-entropy Loss, Focal Loss at least needs to be tuned carefully in order to yield better performance.

As a result, we conclude that common score-based loss functions (i.e. Cross-entropy Loss and Focal Loss) cannot handle different degrees of imbalance without tuning; while our RS Loss can. 

\begin{table}
    \centering
    \setlength{\tabcolsep}{0.3em}
    \caption{RS Loss is robust to class imbalance while score-based loss functions cannot handle imbalanced data in this case. RS Loss successfully trains Faster R-CNN with both relatively balanced (``Random'' sampling) and severely imbalanced (``None'' in the table) data. Numbers in parentheses show positive to negative ratio of sampled examples.}
    \label{tab:SamplingEffect1Score}
    \begin{tabular}{|c|c|c||c|c|c|} \hline
    Loss&RPN&R-CNN&AP&$\mathrm{AP_{50}}$&$\mathrm{AP_{75}}$\\ \hline 
    \multirow{3}{*}{Cross-entropy}&Random&Random&$37.6$&$58.2$&$41.0$ \\ 
    &None&Random&$32.7$&$49.4$&$35.9$\\
    &None&None&$30.1$&$45.3$&$33.2$ \\ \hline
    \multirow{3}{*}{Focal Loss \cite{FocalLoss}}&Random&Random&$31.8$&$47.4$&$35.0$ \\ 
    &None&Random&$32.0$&$47.5$&$35.1$\\
    &None&None&$30.7$&$44.6$&$34.2$ \\ \hline
    \multirow{3}{*}{RS Loss (Ours)}&Random&Random&$38.5$&$58.5$&$41.5$ \\ 
    &None&Random&$39.3$&$\mathbf{59.6}$&$42.3$\\
    &None&None&$\mathbf{39.6}$&$59.5$&$\mathbf{43.0}$ \\ \hline
    \end{tabular}
\end{table}

\subsection{Effect of RS Loss on Efficiency} 
We discuss the effect on efficiency on two levels: (i) training and (ii) inference.

\subsubsection{Effect on Training Efficiency}
Similar to other ranking based loss functions (i.e. AP Loss \cite{APLoss} and aLRP Loss \cite{aLRPLoss}), RS Loss has also quadratic time complexity, and as a result, one training iteration  of RS Loss takes $1.5 \times$ more on average (Table \ref{tab:Time}).

\subsubsection{Effect on Inference Efficiency}
We observed that the methods trained by RS Loss yield larger confidence scores than the baseline methods, which are trained by score-based loss functions (e.g. Cross-entropy Loss). As a result, for inference efficiency, the score threshold to discard detections associated with background before Non-Maximum Suppression (NMS) should be set carefully\footnote{We keep the default settings of the methods in the paper}. Here, we provide examples using multi-stage visual detectors on two datasets:
\begin{itemize}
    \item \textbf{COCO dataset.} Faster R-CNN and Mask R-CNN use $0.05$ as the confidence score threshold on COCO dataset when they are trained by Cross-entropy Loss, that is, all the detections with confidence score less than $0.05$ are regarded as background (i.e. false positive), and they are simply removed from the detection set before NMS. Keeping this setting as $0.05$, RS-R-CNN with ResNet-50 reaches $39.6$ AP and $67.9$ oLRP but with slower inference time than the baseline Cross-entropy Loss,
%at $x.xx$ fps on a single V100 GPU. However, note that this inference time is slower than baseline Faster R-CNN 
which has $21.4$ fps. Then, tuning this score threshold to $0.40$, Faster R-CNN trained by our RS Loss performs exactly the same (see $39.6$ AP and $67.9$ oLRP in Table \ref{tab:Time}) at $22.5$ fps, slightly faster than the baseline Faster R-CNN. Table \ref{tab:Time} presents the results on Faster R-CNN and Mask R-CNN with the tuned confidence score threshold, that is $0.40$. While the performance of models in Table \ref{tab:Time} in terms of oLRP is always equal to the ones with confidence score of $0.05$, in some rare cases we observed negligible performance drop (i.e. up to $0.01$ AP points, e.g. RS Faster R-CNN+ drops from $40.8$ AP to $40.7$ AP).

\item \textbf{LVIS dataset.} Table \ref{tab:LVISconfscores} presents the results of Mask R-CNN on LVIS dataset. Similar to COCO dataset, when we use RS Loss, we prefer a larger confidence score threshold, that is $0.60$, and also we observe that RS Loss is robust to this threshold choice, while the performance of the standard Mask R-CNN degrades rapidly when the score threshold increases. As a result, when the score threshold is set accordingly, our RS-Mask R-CNN yields $25.2$ AP at $11.7$ fps, which outperforms the baseline Mask R-CNN with $21.7$ AP at $11.0$ fps in the best confidence score setting.

\end{itemize}
As a result, the models trained by our RS Loss outputs larger confidence scores, and accordingly, the score threshold needs to be adjusted accordingly for better efficiency.

% We provide in Table \ref{tab:Allresults} the detailed performance (i.e. AP-based, oLRP-based performance measures and fps) of two-stage methods trained with our RS Loss on COCO \textit{minival}.
\begin{table}
    \centering
    \setlength{\tabcolsep}{0.5em}
    \caption{Average iteration time of methods trained by the standard loss vs. RS Loss. On average, training with RS Loss incurs $\sim 1.5 \times$ longer time due to its quadratic time complexity similar to other existing ranking-based loss functions \cite{APLoss,aLRPLoss}. }
    \label{tab:Time}
    \begin{tabular}{|c||c|c|} \hline
    Method&Standard Loss (sec)&RS Loss (sec)\\ \hline
    Faster R-CNN&$0.42$&$0.82$\\ \hline
    Cascade R-CNN&$0.51$&$2.26$\\ \hline
    ATSS &$0.44$&$0.70$\\ \hline
    PAA &$0.57$&$0.99$\\ \hline
    Mask R-CNN&$0.64$&$1.04$\\ \hline
    YOLACT&$0.57$&$0.59$\\ \hline
    SOLOv2-light&$0.64$&$0.90$\\ \hline
    \end{tabular}
\end{table}

\begin{table*}
    \centering
    \caption{Comprehensive performance results of models trained by RS-Loss on COCO \textit{minival}. We report AP-based and oLRP-based performance measures, an also inference time on a single Tesla V100 GPU as fps. We set NMS score threshold of RS-R-CNN and RS-Mask R-CNN to 0.40 for COCO dataset. }
    \setlength{\tabcolsep}{0.4em}
    \label{tab:Allresults}
    \footnotesize
    \begin{tabular}{|c|c|c|c||c|c|c||c|c|c|c|c|} \hline
    Method&Backbone&Epoch&MS train&AP$\uparrow$&$\mathrm{AP_{50}} \uparrow$&$\mathrm{AP_{75}} \uparrow$&oLRP $\downarrow$&$ \mathrm{oLRP_{Loc}} \downarrow$&$ \mathrm{oLRP_{FP}} \downarrow$&$ \mathrm{oLRP_{FN}} \downarrow$ & fps\\ \hline
    \textit{Object Detection}& & & & & & & & & & & \\    
%    RS-Cascade R-CNN&R-50&12& &$41.3$&$58.9$&$44.7$&$66.6$&$?$&$?$&$?$&$?$\\ \hline
    RS-Faster R-CNN&R-50&12&--&$39.6$&$59.5$&$43.0$&$67.9$&$16.3$&$27.8$&$45.4$&$22.5$\\ 
    RS-Mask R-CNN&R-50&12&--&$40.0$&$59.8$&$43.4$&$67.5$&$16.1$&$27.6$&$44.7$&$22.8$\\ 
    RS-Faster R-CNN+&R-50&12&--&$40.7$&$61.4$&$43.8$&$66.9$&$16.3$&$26.4$&$43.7$&$21.5$\\
    RS-Mask R-CNN+&R-50&12&--&$41.1$&$61.4$&$44.8$&$66.6$&$16.0$&$26.2$&$44.0$&$21.1$\\
    RS-Mask R-CNN&R-101&36&$[640,800]$&$44.7$&$64.4$&$48.8$&$63.7$&$14.7$&$24.5$&$41.3$&$17.1$\\ 
    RS-Mask R-CNN+&R-101&36&$[480,960]$&$46.1$&$66.2$&$50.3$&$62.6$&$14.5$&$23.5$&$39.9$&$16.6$\\ 
    RS-Faster R-CNN&R-101-DCN&36&$[480,960]$&$47.6$&$67.8$&$51.2$&$61.1$&$14.4$&$22.7$&$37.9$&$14.1$\\ 
    RS-Faster R-CNN+&R-101-DCN&36&$[480,960]$&$47.6$&$68.1$&$51.9$&$60.9$&$14.7$&$20.9$&$38.2$&$13.6$\\
    RS-Mask R-CNN+&R-101-DCN&36&$[480,960]$&$48.7$&$68.9$&$53.1$&$60.2$&$14.3$&$21.3$&$36.9$&$13.5$\\
    RS-Mask R-CNN+&X-101-DCN&36&$[480,960]$&$49.9$&$70.0$&$54.3$&$59.1$&$14.0$&$20.3$&$36.2$&$6.4$\\
\hline
    \hhline{============}
%    \textit{One-stage Object Detection}& & & & & & & & & & & \\    
%    RS-ATSS&R-50&12& &$?$&$?$&$?$&$?$&$?$&$?$&$?$&$?$\\ 
%    RS-PAA&R-50&12& &$?$&$?$&$?$&$?$&$?$&$?$&$?$&$?$\\ \hline
    \textit{Instance Segmentation}& & & & & & & & & & & \\    
    RS-Mask R-CNN&R-50&12&--&$36.4$&$57.3$&$39.2$&$70.1$&$18.2$&$28.7$&$46.5$&$17.1$\\ 
    RS-Mask R-CNN+&R-50&12&--&$37.3$&$58.6$&$40.2$&$69.4$&$18.1$&$28.0$&$45.3$&$16.7$\\
    RS-Mask R-CNN&R-101&36&$[640,800]$&$40.3$&$61.9$&$43.8$&$66.9$&$17.3$&$24.3$&$43.0$&$14.8$\\ 
    RS-Mask R-CNN+&R-101&36&$[480,960]$&$41.4$&$63.6$&$44.7$&$65.9$&$17.1$&$22.8$&$42.2$&$14.3$\\ 
    RS-Mask R-CNN+&R-101-DCN&36&$[480,960]$&$43.5$&$66.5$&$47.2$&$64.0$&$17.2$&$22.3$&$38.5$&$11.9$\\ 
    RS-Mask R-CNN+&X-101-DCN&36&$[480,960]$&$44.4$&$67.8$&$47.7$&$63.1$&$17.1$&$21.1$&$37.7$&$6.0$\\ 
\hline
%    \textit{One-stage Ins. Segmentation}& & & & & & & & & & & \\    
%    RS-YOLACT&R-50&55& &$?$&$?$&$?$&$?$&$?$&$?$&$?$&$?$\\ 
%    RS-SOLOv2-light&R-34&36&\checkmark&$?$&$?$&$?$&$?$&$?$&$?$&$?$&$?$\\ 
%    RS-SOLOv2&R-101&36&\checkmark&$?$&$?$&$?$&$?$&$?$&$?$&$?$&$?$\\ 
%    \hline
%    RS Cascade R-CNN+&?& & & & \\ \hline
    \end{tabular}
\end{table*}

\begin{table*}
    \centering
    \caption{The performances of RS-Mask R-CNN and baseline Mask R-CNN (i.e. trained by Cross-entropy Loss) over different confidence score thresholds on LVIS v1.0 val set. RS-Mask R-CNN is robust to score threshold while the performance of Mask R-CNN degrades rapidly especially for rare classes. Our best method achieves 25.2 mask AP at 11.7 fps (bold), which is also slightly faster than the best performing method of Mask R-CNN (underlined). Accordingly, we set NMS score threshold of RS-Mask R-CNN to 0.60 for LVIS dataset. Inference time is reported on a single A100 GPU.}
    \label{tab:LVISconfscores}
    \begin{tabular}{|c||c|c|c|c|c|c||c|c|c|c|c|c|} \hline
    \multirow{2}{*}{Score threshold}&\multicolumn{6}{|c|}{Mask R-CNN}&\multicolumn{6}{|c|}{RS-Mask R-CNN}\\ \cline{2-13}
    &$\mathrm{AP_{mask}}$&$\mathrm{AP_{r}}$&$\mathrm{AP_{c}}$&$\mathrm{AP_{f}}$&$\mathrm{AP_{box}}$&fps&$\mathrm{AP_{mask}}$&$\mathrm{AP_{r}}$&$\mathrm{AP_{c}}$&$\mathrm{AP_{f}}$&$\mathrm{AP_{box}}$&fps\\ \hline
    $10^{-4}$&$21.7$&$9.6$&$21.0$&$27.8$&$22.5$&$3.2$&$25.2$&$16.8$&$24.3$&$29.9$&$25.9$&$0.2$\\
    $10^{-3}$&\underline{$21.7$}&\underline{$9.6$}&\underline{$21.0$}&\underline{$27.8$}&\underline{$22.5$}&\underline{$11.0$}&$25.2$&$16.8$&$24.3$&$29.9$&$25.9$&$0.2$\\
    $10^{-2}$&$21.1$&$8.3$&$20.4$&$27.6$&$22.0$&$13.6$&$25.2$&$16.8$&$24.3$&$29.9$&$25.9$&$0.2$\\ 
    $0.10$&$14.7$&$2.0$&$11.6$&$23.6$&$15.4$&$21.4$&$25.2$&$16.8$&$24.3$&$29.9$&$25.9$&$0.2$\\     
    $0.40$&$8.5$&$0.5$&$4.4$&$16.5$&$8.9$&$24.0$&$25.2$&$16.8$&$24.3$&$29.9$&$25.9$&$1.2$\\     
    $0.60$&$6.1$&$0.4$&$2.3$&$12.9$&$6.4$&$24.4$&$\mathbf{25.2}$&$\mathbf{16.8}$&$\mathbf{24.3}$&$\mathbf{29.9}$&$\mathbf{25.9}$&$\mathbf{11.7}$\\     
    $0.80$&$4.0$&$0.1$&$1.1$&$8.8$&$4.1$&$24.8$&$17.0$&$6.3$&$14.3$&$24.7$&$17.6$&$21.2$\\     
    \hline
    \end{tabular}
\end{table*}

%% file: main.bbl
\begin{thebibliography}{10}\itemsep=-1pt

\bibitem{yolact}
Daniel Bolya, Chong Zhou, Fanyi Xiao, and Yong~Jae Lee.
\newblock Yolact: {Real-time} instance segmentation.
\newblock In {\em IEEE/CVF International Conference on Computer Vision (ICCV)},
  2019.

\bibitem{CascadeRCNN}
Zhaowei Cai and Nuno Vasconcelos.
\newblock Cascade {R-CNN:} {D}elving into high quality object detection.
\newblock In {\em IEEE/CVF Conference on Computer Vision and Pattern
  Recognition (CVPR)}, 2018.

\bibitem{D2Det}
Jiale Cao, Hisham Cholakkal, Rao~Muhammad Anwer, Fahad~Shahbaz Khan, Yanwei
  Pang, and Ling Shao.
\newblock D2det: Towards high quality object detection and instance
  segmentation.
\newblock In {\em IEEE/CVF Conference on Computer Vision and Pattern
  Recognition (CVPR)}, 2020.

\bibitem{blendmask}
Hao Chen, Kunyang Sun, Zhi Tian, Chunhua Shen, Yongming Huang, and Youliang
  Yan.
\newblock Blendmask: Top-down meets bottom-up for instance segmentation.
\newblock In {\em IEEE/CVF Conference on Computer Vision and Pattern
  Recognition (CVPR)}, 2020.

\bibitem{SamplingHeuristics}
Joya Chen, Dong Liu, Tong Xu, Shilong Zhang, Shiwei Wu, Bin Luo, Xuezheng Peng,
  and Enhong Chen.
\newblock Is sampling heuristics necessary in training deep object detectors?
\newblock {\em arXiv}, 1909.04868, 2019.

\bibitem{APLoss}
Kean {Chen}, Weiyao {Lin}, Jianguo {li}, John {See}, Ji {Wang}, and Junni
  {Zou}.
\newblock Ap-loss for accurate one-stage object detection.
\newblock {\em IEEE Transactions on Pattern Analysis and Machine Intelligence
  (TPAMI)}, pages 1--1, 2020.

\bibitem{HTC}
Kai Chen, Jiangmiao Pang, Jiaqi Wang, Yu Xiong, Xiaoxiao Li, Shuyang Sun,
  Wansen Feng, Ziwei Liu, Jianping Shi, Wanli Ouyang, Chen~Change Loy, and
  Dahua Lin.
\newblock Hybrid task cascade for instance segmentation.
\newblock In {\em IEEE/CVF Conference on Computer Vision and Pattern
  Recognition (CVPR)}, 2019.

\bibitem{mmdetection}
Kai Chen, Jiaqi Wang, Jiangmiao Pang, Yuhang Cao, Yu Xiong, Xiaoxiao Li,
  Shuyang Sun, Wansen Feng, Ziwei Liu, Jiarui Xu, Zheng Zhang, Dazhi Cheng,
  Chenchen Zhu, Tianheng Cheng, Qijie Zhao, Buyu Li, Xin Lu, Rui Zhu, Yue Wu,
  Jifeng Dai, Jingdong Wang, Jianping Shi, Wanli Ouyang, Chen~Change Loy, and
  Dahua Lin.
\newblock {MMDetection}: Open mmlab detection toolbox and benchmark.
\newblock {\em arXiv}, 1906.07155, 2019.

\bibitem{tensormask}
Xinlei Chen, Ross Girshick, Kaiming He, and Piotr Doll{\'a}r.
\newblock Tensormask: A foundation for dense object segmentation.
\newblock In {\em IEEE/CVF International Conference on Computer Vision (ICCV)},
  2019.

\bibitem{reppointsv2}
Yihong Chen, Zheng Zhang, Yue Cao, Liwei Wang, Stephen Lin, and Han Hu.
\newblock Reppoints v2: Verification meets regression for object detection.
\newblock In {\em Advances in Neural Information Processing Systems (NeurIPS)},
  2020.

\bibitem{LVIS}
Agrim Gupta, Piotr Dollar, and Ross Girshick.
\newblock Lvis: A dataset for large vocabulary instance segmentation.
\newblock In {\em IEEE/CVF Conference on Computer Vision and Pattern
  Recognition (CVPR)}, 2019.

\bibitem{MaskRCNN}
Kaiming He, Georgia Gkioxari, Piotr Dollar, and Ross Girshick.
\newblock {Mask R-CNN}.
\newblock In {\em IEEE/CVF International Conference on Computer Vision (ICCV)},
  2017.

\bibitem{KLLoss}
Yihui He, Chenchen Zhu, Jianren Wang, Marios Savvides, and Xiangyu Zhang.
\newblock Bounding box regression with uncertainty for accurate object
  detection.
\newblock In {\em IEEE/CVF Conference on Computer Vision and Pattern
  Recognition (CVPR)}, 2019.

\bibitem{maskscoring}
Zhaojin Huang, Lichao Huang, Yongchao Gong, Chang Huang, and Xinggang Wang.
\newblock Mask scoring r-cnn.
\newblock In {\em IEEE/CVF Conference on Computer Vision and Pattern
  Recognition (CVPR)}, 2019.

\bibitem{IoUNet}
Borui Jiang, Ruixuan Luo, Jiayuan Mao, Tete Xiao, and Yuning Jiang.
\newblock Acquisition of localization confidence for accurate object detection.
\newblock In {\em The European Conference on Computer Vision (ECCV)}, 2018.

\bibitem{BCNet}
Lei Ke, Yu-Wing Tai, and Chi-Keung Tang.
\newblock Deep occlusion-aware instance segmentation with overlapping bilayers.
\newblock In {\em IEEE/CVF Conference on Computer Vision and Pattern
  Recognition (CVPR)}, 2021.

\bibitem{paa}
Kang Kim and Hee~Seok Lee.
\newblock Probabilistic anchor assignment with iou prediction for object
  detection.
\newblock In {\em The European Conference on Computer Vision (ECCV)}, 2020.

\bibitem{GFL}
Xiang Li, Wenhai Wang, Lijun Wu, Shuo Chen, Xiaolin Hu, Jun Li, Jinhui Tang,
  and Jian Yang.
\newblock Generalized focal loss: Learning qualified and distributed bounding
  boxes for dense object detection.
\newblock In {\em Advances in Neural Information Processing Systems (NeurIPS)},
  2020.

\bibitem{TridentNet}
Yanghao Li, Yuntao Chen, Naiyan Wang, and Zhaoxiang Zhang.
\newblock Scale-aware trident networks for object detection.
\newblock In {\em IEEE/CVF International Conference on Computer Vision (ICCV)},
  2019.

\bibitem{BAGS}
Yu Li, Tao Wang, Bingyi Kang, Sheng Tang, Chunfeng Wang, Jintao Li, and Jiashi
  Feng.
\newblock Overcoming classifier imbalance for long-tail object detection with
  balanced group softmax.
\newblock In {\em IEEE/CVF Conference on Computer Vision and Pattern
  Recognition (CVPR)}, 2020.

\bibitem{FeaturePyramidNetwork}
Tsung{-}Yi Lin, Piotr Doll{\'{a}}r, Ross~B. Girshick, Kaiming He, Bharath
  Hariharan, and Serge~J. Belongie.
\newblock Feature pyramid networks for object detection.
\newblock In {\em IEEE/CVF Conference on Computer Vision and Pattern
  Recognition (CVPR)}, 2017.

\bibitem{FocalLoss}
Tsung-Yi {Lin}, Priya {Goyal}, Ross {Girshick}, Kaiming {He}, and Piotr
  {Dollár}.
\newblock Focal loss for dense object detection.
\newblock {\em IEEE Transactions on Pattern Analysis and Machine Intelligence
  (TPAMI)}, 42(2):318--327, 2020.

\bibitem{COCO}
Tsung-Yi Lin, Michael Maire, Serge Belongie, James Hays, Pietro Perona, Deva
  Ramanan, Piotr Doll{\'{a}}r, and C~Lawrence Zitnick.
\newblock {Microsoft COCO: Common Objects in Context}.
\newblock In {\em The European Conference on Computer Vision (ECCV)}, 2014.

\bibitem{autoloss}
Peidong Liu, Gengwei Zhang, Bochao Wang, Hang Xu, Xiaodan Liang, Yong Jiang,
  and Zhenguo Li.
\newblock Loss function discovery for object detection via
  convergence-simulation driven search.
\newblock In {\em International Conference on Learning Representations (ICLR)},
  2021.

\bibitem{SSD}
Wei Liu, Dragomir Anguelov, Dumitru Erhan, Christian Szegedy, Scott~E. Reed,
  Cheng{-}Yang Fu, and Alexander~C. Berg.
\newblock {SSD:} single shot multibox detector.
\newblock In {\em The European Conference on Computer Vision (ECCV)}, 2016.

\bibitem{LRP}
Kemal Oksuz, Baris~Can Cam, Emre Akbas, and Sinan Kalkan.
\newblock Localization recall precision {(LRP)}: A new performance metric for
  object detection.
\newblock In {\em The European Conference on Computer Vision (ECCV)}, 2018.

\bibitem{LRParXiv}
Kemal Oksuz, Baris~Can Cam, Emre Akbas, and Sinan Kalkan.
\newblock One metric to measure them all: Localisation recall precision (lrp)
  for evaluating visual detection tasks.
\newblock {\em arXiv}, 2011.10772, 2020.

\bibitem{aLRPLoss}
Kemal Oksuz, Baris~Can Cam, Emre Akbas, and Sinan Kalkan.
\newblock A ranking-based, balanced loss function unifying classification and
  localisation in object detection.
\newblock In {\em Advances in Neural Information Processing Systems (NeurIPS)},
  2020.

\bibitem{Review}
Kemal {Oksuz}, Baris~Can {Cam}, Sinan {Kalkan}, and Emre {Akbas}.
\newblock Imbalance problems in object detection: A review.
\newblock {\em IEEE Transactions on Pattern Analysis and Machine Intelligence
  (TPAMI)}, pages 1--1, 2020.

\bibitem{LibraRCNN}
Jiangmiao Pang, Kai Chen, Jianping Shi, Huajun Feng, Wanli Ouyang, and Dahua
  Lin.
\newblock Libra {R-CNN:} {T}owards balanced learning for object detection.
\newblock In {\em IEEE/CVF Conference on Computer Vision and Pattern
  Recognition (CVPR)}, 2019.

\bibitem{DRLoss}
Qi {Qian}, Lei {Chen}, Hao {Li}, and Rong {Jin}.
\newblock Dr loss: Improving object detection by distributional ranking.
\newblock In {\em IEEE/CVF Conference on Computer Vision and Pattern
  Recognition (CVPR)}, 2020.

\bibitem{FasterRCNN}
Shaoqing Ren, Kaiming He, Ross Girshick, and Jian Sun.
\newblock Faster {R-CNN:} {T}owards real-time object detection with region
  proposal networks.
\newblock {\em IEEE Transactions on Pattern Analysis and Machine Intelligence
  (TPAMI)}, 39(6):1137--1149, 2017.

\bibitem{GIoULoss}
Hamid Rezatofighi, Nathan Tsoi, JunYoung Gwak, Amir Sadeghian, Ian Reid, and
  Silvio Savarese.
\newblock Generalized intersection over union: A metric and a loss for bounding
  box regression.
\newblock In {\em IEEE/CVF Conference on Computer Vision and Pattern
  Recognition (CVPR)}, 2019.

\bibitem{RankBasedBlackboxDifferentiation}
Michal Rolínek, Vít Musil, Anselm Paulus, Marin Vlastelica, Claudio
  Michaelis, and Georg Martius.
\newblock Optimizing rank-based metrics with blackbox differentiation.
\newblock In {\em IEEE/CVF Conference on Computer Vision and Pattern
  Recognition (CVPR)}, 2020.

\bibitem{Rosenblatt}
F. Rosenblatt.
\newblock The perceptron: A probabilistic model for information storage and
  organization in the brain.
\newblock {\em Psychological Review}, pages 65--386, 1958.

\bibitem{OHEM}
Abhinav Shrivastava, Abhinav Gupta, and Ross Girshick.
\newblock Training region-based object detectors with online hard example
  mining.
\newblock In {\em IEEE/CVF Conference on Computer Vision and Pattern
  Recognition (CVPR)}, 2016.

\bibitem{Eqv2}
Jingru Tan, Xin Lu, Gang Zhang, Changqing Yin, and Quanquan Li.
\newblock Equalization loss v2: A new gradient balance approach for long-tailed
  object detection.
\newblock In {\em IEEE/CVF Conference on Computer Vision and Pattern
  Recognition (CVPR)}, 2021.

\bibitem{FCOS}
Zhi Tian, Chunhua Shen, Hao Chen, and Tong He.
\newblock Fcos: Fully convolutional one-stage object detection.
\newblock In {\em IEEE/CVF International Conference on Computer Vision (ICCV)},
  2019.

\bibitem{carafe}
Jiaqi Wang, Kai Chen, Rui Xu, Ziwei Liu, Chen~Change Loy, and Dahua Lin.
\newblock Carafe: Content-aware reassembly of features.
\newblock In {\em IEEE/CVF International Conference on Computer Vision (ICCV)},
  2019.

\bibitem{solov2}
Xinlong Wang, Rufeng Zhang, Tao Kong, Lei Li, and Chunhua Shen.
\newblock Solov2: Dynamic and fast instance segmentation.
\newblock In {\em Advances in Neural Information Processing Systems (NeurIPS)},
  2020.

\bibitem{centermask}
Yuqing Wang, Zhaoliang Xu, Hao Shen, Baoshan Cheng, and Lirong Yang.
\newblock Centermask: Single shot instance segmentation with point
  representation.
\newblock In {\em IEEE/CVF Conference on Computer Vision and Pattern
  Recognition (CVPR)}, 2020.

\bibitem{polarmask}
Enze Xie, Peize Sun, Xiaoge Song, Wenhai Wang, Xuebo Liu, Ding Liang, Chunhua
  Shen, and Ping Luo.
\newblock Polarmask: Single shot instance segmentation with polar
  representation.
\newblock In {\em IEEE/CVF Conference on Computer Vision and Pattern
  Recognition (CVPR)}, 2020.

\bibitem{dynamicrcnn}
Hongkai Zhang, Hong Chang, Bingpeng Ma, Naiyan Wang, and Xilin Chen.
\newblock Dynamic r-cnn: Towards high quality object detection via dynamic
  training.
\newblock In {\em The European Conference on Computer Vision (ECCV)}, 2020.

\bibitem{ATSS}
Shifeng Zhang, Cheng Chi, Yongqiang Yao, Zhen Lei, and Stan~Z. Li.
\newblock Bridging the gap between anchor-based and anchor-free detection via
  adaptive training sample selection.
\newblock In {\em IEEE/CVF Conference on Computer Vision and Pattern
  Recognition (CVPR)}, 2020.

\bibitem{DCNv2}
Xizhou Zhu, Han Hu, Stephen Lin, and Jifeng Dai.
\newblock Deformable convnets v2: More deformable, better results.
\newblock In {\em IEEE/CVF Conference on Computer Vision and Pattern
  Recognition (CVPR)}, 2019.

\end{thebibliography}
